\documentclass{article} 
\usepackage{iclr2019_conference,times}

\usepackage{amsmath,amsfonts,bm}
\usepackage{ dsfont }

\def\eqref#1{equation~\ref{#1}}

\def\1{\bm{1}}

\DeclareMathAlphabet{\mathsfit}{\encodingdefault}{\sfdefault}{m}{sl}
\SetMathAlphabet{\mathsfit}{bold}{\encodingdefault}{\sfdefault}{bx}{n}

\newcommand{\R}{\mathbb{R}}

\usepackage[utf8]{inputenc} 
\usepackage[T1]{fontenc}    
\usepackage{hyperref}       
\usepackage{url}            
\usepackage{booktabs}       
\usepackage{amsfonts}       
\usepackage{nicefrac}       
\usepackage{microtype}      
\usepackage{amsmath,graphicx}
\usepackage{amssymb}
\usepackage{amsthm}
\usepackage{enumitem}
\usepackage{overpic}
\usepackage{color}
\usepackage{tikz}
\usepackage{adjustbox}
\usepackage{enumerate}
\usepackage{wrapfig}
\usepackage{lipsum}
\usepackage{hyperref}
\usepackage{url}
\usepackage{tikz}
\usepackage{adjustbox}
\usepackage{subfig}
\usepackage{enumerate}
\usepackage{epstopdf}

\newtheorem{theorem}{Theorem}[section]
\newtheorem{lemma}[theorem]{Lemma}
\newtheorem{corollary}[theorem]{Corollary}

\newcommand{\expe}{\mathbb{E}}

\title{Supervised Community Detection with Line Graph Neural Networks}

\author{Zhengdao Chen$^{a}$, Lisha Li$^{b}$,
Joan Bruna$^{ac}$\thanks{Corresponding emails: \href{mailto:zc1216@nyu.edu}{zc1216@nyu.edu} and \href{mailto:bruna@cims.nyu.edu}{bruna@cims.nyu.edu}.}\\[1ex]
   \hspace{3em}\parbox{0.9\linewidth}{
     $^a$ Courant Institute of Mathematical Sciences, New York University, New York, NY\\
     $^b$ Rosebud AI, California \\
     $^c$ Center for Data Science, New York University, New York, NY}
     }

\iclrfinalcopy 
\begin{document}

\maketitle

\begin{abstract}
Traditionally, community detection in graphs can be solved using spectral methods or posterior inference under probabilistic graphical models. Focusing on random graph families such as the stochastic block model, recent research has unified both approaches and identified both statistical and computational detection thresholds in terms of the signal-to-noise ratio. By recasting community detection as a node-wise classification problem on graphs, we can also study it from a learning perspective.
We present a novel family of Graph Neural Networks (GNNs) for solving community detection problems in a supervised learning setting. We show that, in a data-driven manner and without access to the underlying generative models, they can match or even surpass the performance of the belief propagation algorithm on binary and multi-class stochastic block models, which is believed to reach the computational threshold. In particular, we propose to augment GNNs with the non-backtracking operator defined on the line graph of edge adjacencies. Our models also achieve good performance on real-world datasets.  In addition, we perform the first analysis of the optimization landscape of training linear GNNs for community detection problems, demonstrating that under certain simplifications and assumptions, the loss values at local and global minima are not far apart.
\end{abstract}

\section{Introduction}



Graph inference problems encompass a large class of tasks and domains, from posterior inference in probabilistic graphical models to community detection and ranking in generic networks, image segmentation or compressed sensing on non-Euclidean domains. They are motivated both by practical applications, such as in the case of PageRank \citep{ilprints422}, and also by fundamental questions on the algorithmic hardness of solving such tasks. 

From a data-driven perspective, these problems can be formulated in supervised, semi-supervised and unsupervised learning settings. In the supervised case, one assumes a dataset of graphs with labels on their nodes, edges or the entire graphs, and attempts to perform node-wise, edge-wise and graph-wise classification by optimizing a loss over a certain parametric class, e.g. neural networks. Graph Neural Networks (GNNs)
are natural extensions of Convolutional Neural Networks (CNN) to graph-structured data, and have emerged as a powerful class of algorithms to perform complex graph inference leveraging labeled data \citep{gori2005new, bruna2013spectral, duvenaud2015convolutional, bronstein2016geometric, kipf2016semi, defferrard2016convolutional, hamilton2017inductive}. In essence, these neural networks learn cascaded linear combinations of intrinsic graph operators interleaved with node-wise (or edge-wise) activation functions. Since they utilize intrinsic graph operators, they can be applied to varying input graphs, and they offer the same parameter sharing advantages as their CNN counterparts.

In this work, we focus on community detection problems, a wide class of node classification tasks that attempt to discover a clustered, segmented structure within a graph. The traditional algorithmic approaches to this problem include a rich class of spectral methods, which take advantage of the spectrum of certain operators defined on the graph, as well as approximate message-passing methods such as belief propagation (BP), which performs approximate posterior inference under predefined graphical models \citep{decelle2011asymptotic}. Focusing on the supervised setting, we study the ability of GNNs to approximate, generalize and even improve upon these class of algorithms. Our motivation is two-fold. On the one hand, this problem exhibits algorithmic hardness on some settings, opening up the possibility to discover more efficient algorithms than the current ones. On the other hand, many practical scenarios fall beyond pre-specified probabilistic models, hence calling for data-driven solutions. 

We propose modifications to the GNN architecture, which allow it to exploit edge adjacency information, by incorporating the non-backtracking operator of the graph. This operator is defined over the edges of the graph and allows a directed flow of information even when the original graph is undirected. It was introduced to community detection problems by \cite{krzakala2013spectral}, who propose a spectral method based on the non-backtracking operator.  We refer to the resulting GNN model as a \emph{Line Graph Neural Network (LGNN)}.
Focusing on important random graph families exhibiting community structure, such as the stochastic block model (SBM) and the geometric block model (GBM), we demonstrate improvements in the performance by our GNN and LGNN models compared to other methods including spectral methods and BP in regimes within the computational-to-statistical gap \citep{abbe2017community}. In fact, some gains can already be obtained with \emph{linear} LGNNs, which can be interpreted as data-driven versions of power iteration algorithms.

Besides community detection tasks, GNN and LGNN can be applied to other node-wise classification problems too. The reason we are focusing on community detection is that it has a rich theoretical literature where different algorithms have been proposed and fundamental limits in terms of computational and statistical (or information-theoretic) thresholds have been established in several scenarios. Moreover, synthetic datasets can be easily generated for community detection tasks. Therefore, besides the practical value of community detection, we think it is a nice platform for comparing against traditional non-data-driven algorithms.

The good performances of GNN and LGNN motivate our second main contribution: an analysis of the optimization landscape of simplified linear GNN models when trained under a given graph distribution. Under reparametrization, we provide an upper bound on the \emph{energy gap} controlling the loss difference between local and global minima. With some assumptions on the spectral concentration of certain random matrices, this energy gap will shrink as the size of the input graphs increases, which would mean that the optimization landscape is benign on large enough graphs.


\paragraph{Summary of Main Contributions:}
\begin{itemize}
\item We define a GNN model based on a family of multiscale graph operators, and
propose to augment it using the line graph and the non-backtracking operator, which yields improvements in supervised community detection tasks. 
\item For graphs generated from stochastic block models (SBMs), our models reach detection thresholds in a purely data-driven fashion, outperforming belief propagation (BP) in hard regimes. Our models also succeed on graphs from the geometric block model (GBM).
\item Our models perform well in detecting communities in real datasets from SNAP.
\item We perform the first analysis of the learning landscape of linear GNN models, showing that under certain simplifications and assumptions, the local minima are confined in low-loss regions. 
\end{itemize}

\section{Problem setup}
\label{setupsect}



Community detection is a specific type of node-classification tasks in which 
given an input graph $G = (V,E)$, we want to predict an underlying labeling function $y: V \to \{1, \dots, C\}$ that encodes a partition of $V$ into $C$ communities. 
We consider the supervised learning setting, where a training set $\{ (G_t, y_t) \}_{t \leq T}$ is given, with which we train a model that predicts $\hat{y} = \Phi_{\theta}(G)$ by minimizing a loss function of the form
$$L(\theta) = \frac{1}{T} \sum_{t \leq T} \ell( \Phi_{\theta}(G_t), y_t)~$$
\vspace{-0.4cm}

Since $y$ encodes a partition of $C$ groups, the specific label of each node is only important up to a global permutation of $\{1, \dots, C\}$. Section \ref{permutationlosssec} describes how to construct loss functions $\ell$ with such a property. 
Moreover, a permutation of the node indices translates into the same permutation applied to the labels, which justifies using models $\Phi$ that are equivariant to node permutations. Also, we are interested in inferring properties of community detection algorithms that do not depend on the specific size of the graphs, and therefore require that the model $\Phi$ accepts graphs of variable size for the same set of parameters. We also assume that $C$ is known.




\section{Related works}
\label{relatedsect}

GNN was first proposed in \cite{gori2005new, GNN}. \cite{Bruna} generalize convolutional neural networks on general undirected graphs by using the graph Laplacian's eigenbasis. This was the first time the Laplacian operator was used in a neural network architecture to perform classification on graph inputs. \cite{defferrard2016convolutional} consider a symmetric Laplacian generator to define a multiscale GNN architecture, demonstrated on classification tasks. Similarly, \cite{kipf2016semi} use a similar generator as effective embedding mechanisms for graph signals and applies it to semi-supervised tasks. 
This is the closest application of GNNs to our current contribution. However, we highlight that semi-supervised 
learning requires bootstrapping the estimation with a subset of labeled nodes, and is mainly interested in \textit{transductive} learning within a single, fixed graph. In comparison, our setup considers \textit{inductive} community detection across a distribution of input graphs and assumes no initial labeling on the graphs in the test dataset except for the adjacency information.

There have been several extensions of GNNs by modifying their non-linear activation functions, parameter sharing strategies, and choice of graph operators \citep{li2015gated, sukhbaatar2016learning,duvenaud2015convolutional,niepert2016learning}.
In particular, \cite{gilmer2017neural} interpret the GNN architecture as learning an approximate message-passing algorithm, which extends the 
learning of hidden representations to graph edges in addition to graph nodes. 
Recently, \cite{velickovic2017graph} relate adjacency learning with attention mechanisms, 
and \cite{vaswani2017attention} propose a similar architecture in the context of machine translation. 
Another recent and related piece of work is by \cite{kondor2018covariant}, who propose 
a generalization of GNN that captures high-order node interactions through covariant tensor algebra. Our approach to extend the expressive power of GNN using the line graph 
may be seen as an alternative to capture such high-order interactions. 

Our energy landscape analysis is related to the recent paper by \cite{shamir2018resnets}, which establishes an energy bound on the local minima arising in the optimization of ResNets. In our case, we exploit the properties of the community detection problem to produce an energy bound that depends on the concentration of certain random matrices, which one may hope for as the size of the input graphs increases.  
Finally, \cite{zhang2016robust}'s work on data regularization for clustering and rank estimation is also motivated by the success of using Bethe-Hessian-like perturbations to improve spectral methods on sparse networks. It finds good perturbations via matrix perturbations and also has successes on the stochastic block model. 







\section{Line Graph Neural Networks}
\label{gnnsec}

\begin{figure}[ht]
    \centering
    {{\includegraphics[width=0.9\textwidth]{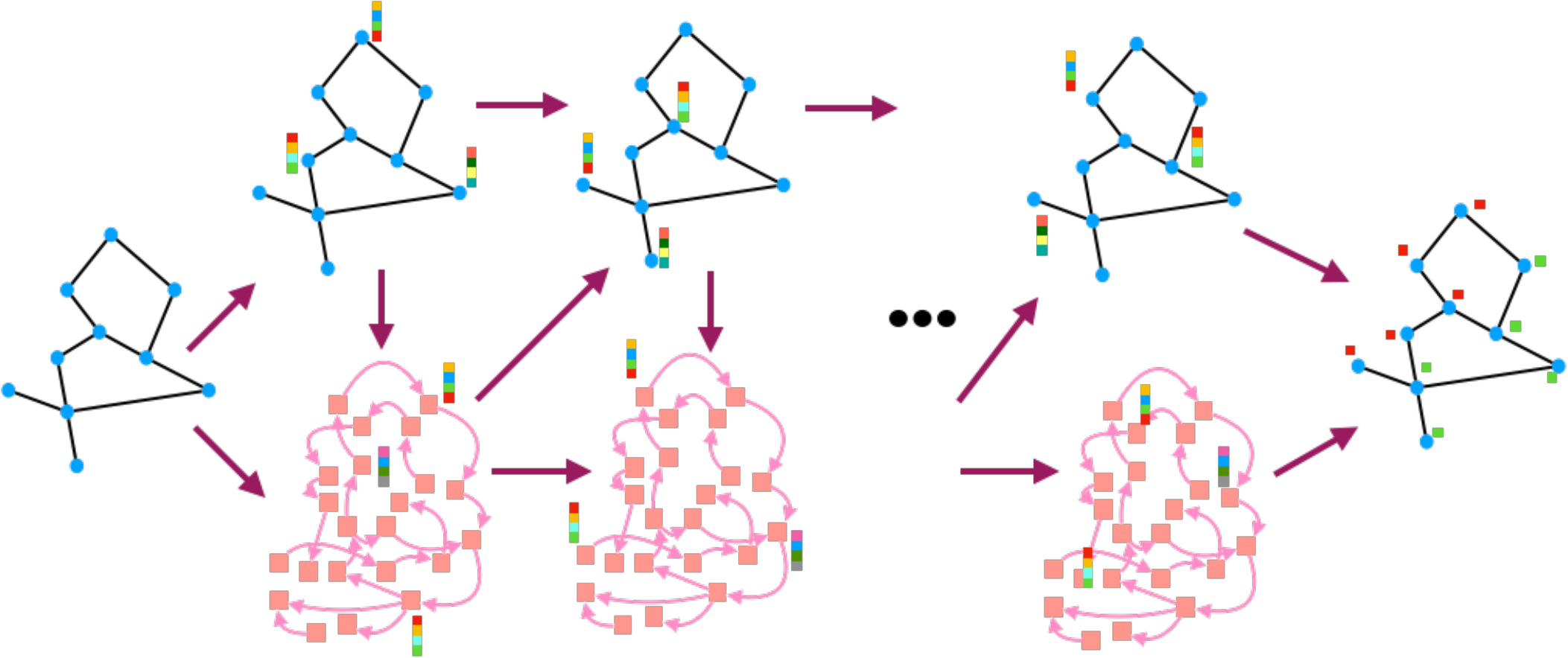}}}
        \caption{Overview of the architecture of LGNN (Section \ref{linegraphgnn}). Given a graph $G$, we construct its line graph $L(G)$ with the non-backtracking operator (Figure \ref{fig:linegraph1}). In every layer, the states of all nodes in $G$ and $L(G)$ are updated according to (\ref{gnneqline}). The final states of nodes in G are used to predict node-wise labels, and the trainining is performed end-to-end using standard backpropagation with a label permutation invariant loss (Section \ref{permutationlosssec}).}
    \label{fig:model}
\end{figure}

This section introduces our GNN architectures that include the power graph adjacency (Section \ref{GNNbasic}) and 
its extension to line graphs using the non-backtracking operator (Section \ref{linegraphgnn}), as 
well as the design of losses invariant to global label permutations (Section \ref{permutationlosssec}).

\subsection{Graph Neural Networks using a family of multiscale graph operators}
\label{GNNbasic}
Given a graph $G = (V, E)$ and a vector $x \in \mathbb{R}^{|V| \times b}$ of node features, we consider intrinsic linear operators of the graph that act locally on $x$, which can be represented as $|V|$-by-$|V|$ matrices. For example, the \textit{adjacency matrix} $A$ is defined entry-wise by $A_{i_1 i_2} = 1$ if $(i_1, i_2) \in E$ and $A_{i_1 i_2} = 0$ if $(i_1, i_2) \notin E$, for every pair $(i_1, i_2) \in V \times V$. The \textit{degree matrix} $D$ is a diagonal matrix with $D_{ii}$ being the degree of the $i$th node, and it can be expressed as $\text{diag}( A {\mathds 1})$. We can also define \textit{power graph adjacency matrices} as $A_{J} = \min(1, A^{2^J})$, which encodes $2^J$-hop neighborhoods into a binary graph for $J \in \mathbb{N}^*$. Finally, there is also the \textit{identity matrix}, $I$. 

Having a family of such matrices, $\mathcal{F} = \{I, D, A, A_{2}, ..., A_{J} \}$ with a certain $J$,
we can define a multiscale GNN layer that maps $x^{(k)} \in \R^{|V| \times b_k}$ 
to $x^{(k+1)} \in \R^{|V| \times b_{k+1}}$ as follows. First, we compute
\begin{equation}
\label{gnneq}
{z^{(k+1)}} = \rho \left[ \sum_{O_i \in \mathcal{F}} O_i x^{(k)} \theta_i \right], \hspace{10pt}
{\overline{z}^{(k+1)}} = \sum_{O_i \in \mathcal{F}} O_i x^{(k)} \theta_i
\end{equation}
where $\theta_j \in \mathbb{R}^{b_k \times \frac{b_{k+1}}{2}}$ are trainable parameters and $\rho(\cdot)$ is a point-wise nonlinear activation function, chosen in this work to be the ReLU function, i.e. $\rho(z) = \max(0,z)$ for $z \in \mathbb{R}$. Then we define $ x^{(k+1)} = [z^{(k+1)}, \overline{z}^{(k+1)}] \in  \mathbb{R}^{|V| \times b_{k+1}}$
as the concatenation of $z^{(k+1)}$ and $\overline{z}^{(k+1)}$.
The layer thus includes linear skip connections via $\overline{z}^{(k)}$, both to ease with the optimization when using large number of layers (similar to residual connections \citep{he2016deep}) and to increase the expressive power of the model by enabling it to perform power iterations. Since the spectral radius of the learned linear operators in (\ref{gnneq}) can grow as the optimization progresses, the cascade of GNN layers can become unstable to training. 
In order to mitigate this effect, we perform instance normalization (or spatial batch normalization with one graph per batch) \citep{ioffe2015batch, ulyanov2016instance} at each layer.
The initial node states $x^{(0)}$ are set to be the node attributes if they are present in the data, and otherwise the degrees of the nodes, i.e., $x^{(0)}_i = D_{ii}$. 

Note that the model $\Phi(G, x^{(0)})=x^{(K)}$ satisfies the permutation equivariance property required 
for node classification: given a permutation $\pi$ among the nodes in the graph, $\Phi(\pi \circ G, \Pi x^{(0)}) = \Pi \Phi(G, x^{(0)})$, where $\Pi$ is the $|V| \times |V|$ permutation matrix associated with $\pi$. 

\paragraph{Analogy with power iterations}

In our setup, instance normalization not only prevents gradient blowup, but also performs the orthogonalisation relative to the 
constant vector, which reinforces the analogy with the spectral methods for community detection, some background of which is described in Appendix \ref{laplaciansetup}. In short, under certain conditions, the community structure of the graph is correlated with both the eigenvector of $A$ corresponding to its second largest eigenvalue and the eigenvector of the \textit{Laplacian matrix}, $L = D - A$, corresponding to its second smallest eigenvalue (the latter often called the \textit{Fiedler vector}). Thus, spectral methods for community detection performs power iterations on these matrices to obtain the eigenvectors of interest and predicts the community structure based on them. 
For example, to extract the Fiedler vector, after finding the eigenvector $v$ corresponding to the smallest eigenvalue of $L$, one can then perform projected power iterations on $\tilde{L} := \| L \| I - L$ 
by iteratively computing $y^{(n+1)} = \tilde{L} x^{(n)}$ and $x^{(n+1)} = \frac{y^{(n+1)} - v^T v y^{(n+1)} }{\| y^{(n+1)} - v^T v y^{(n+1)} \|}~.$
As $v$ is in fact a constant vector, the normalization here is analogous to the instance normalization step in the GNN layer defined above.

As explained in Appendix \ref{laplaciansetup}, the graph Laplacian is not ideal for spectral clustering to operate well in the sparse regime as compared to the Bethe Hessian matrix, which explores the space of matrices generated by $\{I, D, A\}$, just like our GNN model. Moreover, the expressive power of our GNN is further increased by adding multiscale versions of $A$. 
We can choose the depth of the GNN to be of the order of the graph diameter, so that all nodes obtain information from the entire graph. In sparse graphs with small diameter, this architecture offers excellent scalability and computational complexity. 
Indeed, in many social networks diameters are constant (due to hubs) or  $\log (|V|)$, as in the stochastic block model in the constant or $\log(|V|)$ average degree regime \citep{riordan2010diameter}. This results in a model with computational complexity on the order of $|V| \log(|V|) $, making it amenable to large-scale graphs.



\subsection{LGNN: GNN on line graphs with the non-backtracking operator}
\label{linegraphgnn}
Belief propagation (BP) is a dynamical-programming-style algorithm for computing exactly or approximating marginal distributions in graphical model inferences \citep{pearl1982reverend, yedidia2003understanding}. 
BP operates by passing messages iteratively on the non-backtracking edge adjacency structure, which we introduce in details in Appendix \ref{bpsection}.
In this section, we describe an upgraded GNN model that exploits the non-backtracking structure, which can be viewed as a data-driven generalization of BP.

\begin{figure}
    \centering
    \includegraphics[width=0.25\linewidth]{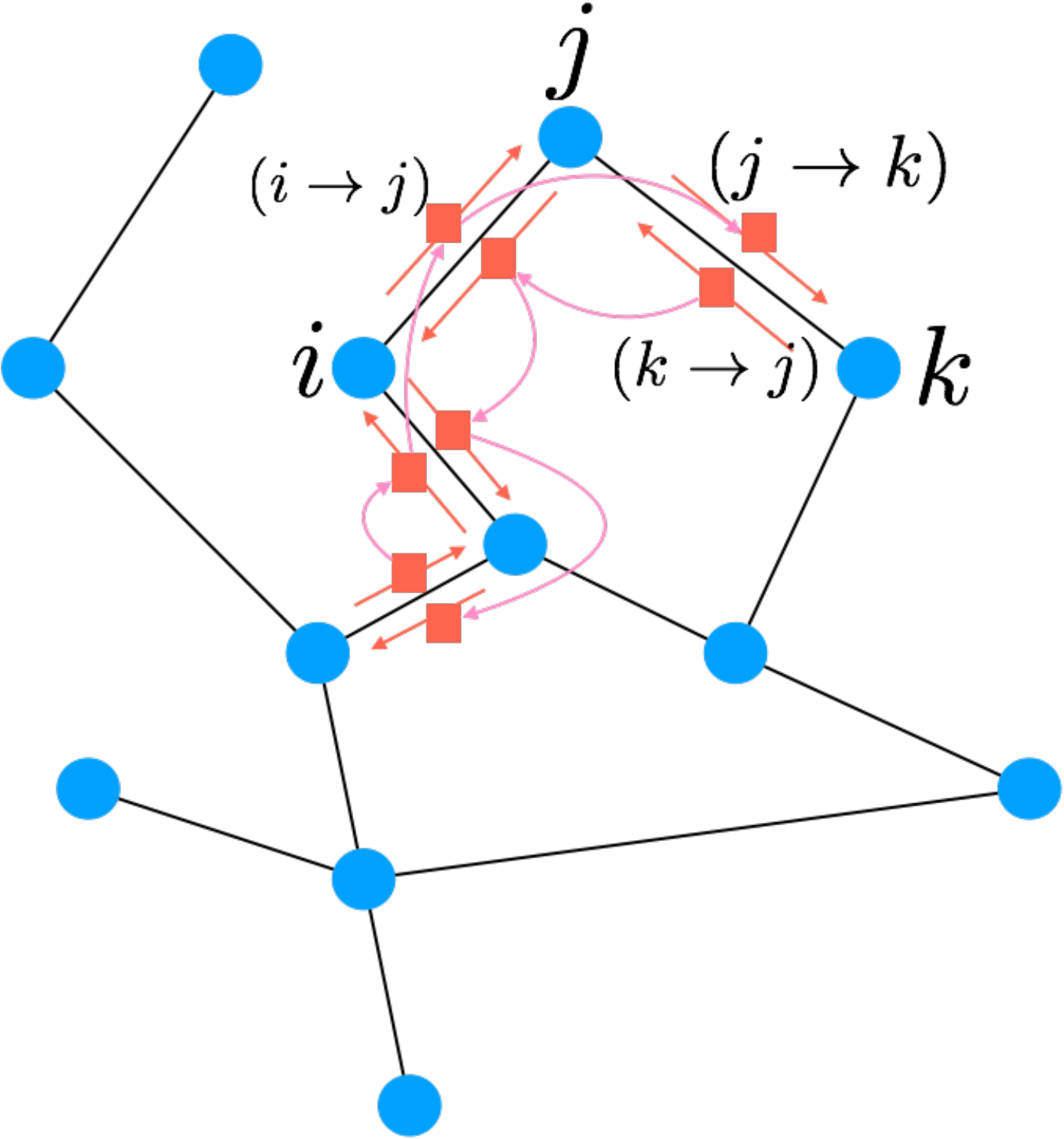}
     \caption{Construction of the line graph $L(G)$ using the non-backtracking matrix. The nodes of $L(G)$ correspond to oriented edges of $G$.}
    \label{fig:linegraph1}
\end{figure}

Given an undirected graph $G=(V,E)$, its line graph $L(G)=(V_L, E_L)$ encodes the \textit{directed} edge adjacency structure of $G$. The vertices of $L(G)$ consist of the ordered edges in $E$, 
i.e., 
$V_L = \{ (i \to j): \, (i,j) \in E\}$, which means that 
$|V_L| = 2 |E|$.
The edge set $E_L$ of $L(G)$ is given by the non-backtracking matrix $B \in \R^{2|E| \times 2|E|}$ defined as $$B_{(i\to j), (i'\to j')}=\left\{ 
\begin{array}{cc}
1 & \text{if } j=i' \text{ and } j'\neq i\,,\\
0 & \text{otherwise.}
\end{array} \right.$$
\vspace{-0.25cm}

This matrix enables the directed propagation of information on the line graph, and was first proposed in the context of
community detection on sparse graphs in \cite{krzakala2013spectral}.
The message-passing rules of BP can be expressed as a diffusion in the line graph $L(G)$ using 
this non-backtracking operator, with specific choices of activation function that turn product of beliefs into sums. Detailed explanations are given in Appendix \ref{app.spectral}.

Thus, a natural extension of the GNN architecture presented in Section \ref{GNNbasic} is to 
consider a second GNN defined on $L(G)$, where $B$ and $D_{B}= \mathrm{diag}(B \mathds{1})$ play the role of the adjacency and the degree matrices, respectively. Analogous to $A_J$, we also define $B_J= \min(1, B^{2^J})$. These operators allow us to consider edge states that update according 
to the edge adjacency of $G$. Moreover, edge and node states communicate at each layer using the \textit{unsigned} and \textit{signed incidence matrices} $P, \tilde{P} \in \{0,1\}^{|V| \times 2|E| }$, defined as 
$P_{i, (i\to j)} = 1$, $P_{j, (i\to j)} = 1$, $\tilde{P}_{i, (i\to j)} = 1$, $\tilde{P}_{j, (i\to j)} = -1$ and $0$ otherwise. Together with skip linear connections $\overline{z}^{(k+1)}$ and $\overline{w}^{(k+1)}$ defined in a way analogous to (\ref{gnneq}), the update rule at each layer can be written as
\begin{equation}
\label{gnneqline}
\begin{aligned}
z^{(k+1)} &= \rho \left[ \sum_{O_i \in \mathcal{F}} O_i x^{(k)} \theta_i + \sum_{O''_j \in \mathcal{F}''} O''_j y^{(k)} \theta''_i \right] \\
w^{(k+1)} &= \rho \left[ \sum_{O'_l \in \mathcal{F}'} O'_l y^{(k)} \theta'_i + \sum_{O''_j \in \mathcal{F}''} (O''_j)^T x^{(k+1)} \theta'''_j \right]
\end{aligned}
\end{equation}
where $\mathcal{F} = \{ I, D, A, A_{2}, \dots, A_{J}\}$, $\mathcal{F}' = \{I_B, D_B, B, B_{2}, \dots, B_{J}\}$, $\mathcal{F}'' = \{P, \tilde{P}\}$, and the trainable parameters are $\theta_i, \theta'_i, \theta''_i \in \mathbb{R}^{b_k \times b_{k+1}}$ and $\theta'''_i \in \mathbb{R}^{b_{k+1} \times b_{k+1}}$. We call such a model a \textit{Line Graph Neural Network (LGNN)}. 


In our experiments, we set $x^{(0)} = \mathrm{deg}(A)$ and $y^{(0)} = \mathrm{deg}(B)$. For graph families whose average degree $\overline{d}$ remains constant as $|V|$ grows, the line graph has size of $O(\overline{d} |V|)$, 
and therefore the model is feasible computationally. Furthermore, the construction of line graphs can be iterated to generate $L(L(G))$, $L(L(L(G)))$, etc. to yield a \emph{line graph hierarchy}, which can capture higher-order interactions among nodes of $G$. Such a hierarchical construction is related to other recent efforts to generalize GNNs \citep{kondor2018covariant, morris2019weisfeiler}. 

\paragraph{Learning directed edge features from an undirected graph} 
Several authors have proposed to combine node and edge feature learning, such as \citep{battaglia2016interaction, kearnes2016molecular, gilmer2017neural, velickovic2017graph}.
However, we are not aware of works that consider the edge adjacency structure provided by the non-backtracking matrix on the line graph. 
With non-backtracking matrix, our LGNN can be interpreted as learning \emph{directed} edge features from an \emph{undirected} graph. Indeed, if each node $i$ contains two distinct sets of features $x_s(i)$ and $x_r(i)$, the non-backtracking operator constructs edge features from node features while preserving orientation: For an edge $e=(i,j)$, our model is equivalent to constructing oriented edge features $f_{i\to j} = g(x_s(i), x_r(j))$ and $f_{j \to i} = g(x_r(i), x_s(j))$ (where $g$ is trainable and not necessarily commutative on its arguments) that are subsequently propagated through the graph. 
To demonstrate the benefit of incorporating such local oriented structures, we will compare LGNN with a modified version, \textit{symmetric LGNN (LGNN-S)}. LGNN-S is based on an alternative line graph of size $|E|$ whose nodes are the \textit{undirected} edges of the original graph, and where two such undirected edges of $G$ are adjacent if and only if they share one common node in $G$; also, we set $\mathcal{F}'' = \{ P \}$ in LGNN-S, with $P \in \mathbb{R}^{|V| \times |E|}$ defined as $P_{i, (j, k)} = 1$ if $i = j$ or $k$ and $0$ otherwise. In addition, we also define \textit{linear LGNN (LGNN-L)} as the LGNN that drops the nonlinear activation functions $\rho$ in (\ref{gnneqline}).
\subsection{A loss function invariant under label permutation}
\label{permutationlosssec}

Let $G=(V,E)$ be the input graph and $y_i$ be the ground truth 
community label of node $i$. Let $[C]:=\{1, \dots, C\}$ denote the set of all community labels, and consider first the case where 
communities do not overlap.
After applying the softmax function at the end of the model, for each $c \in [C]$, we interpret the $c$th dimension of the model's output at node $i$ as the conditional probability that the node belongs to community $c$: $o_{i, c} = p(y_i = c~|\theta,G)$. 
Since the community structure is defined up to global permutations of the labels, we define the loss function as
\begin{equation}
\label{lossfunction}
\ell( \theta ) =  \min_{ \pi \in S_{C}} - \sum_{i \in V} \log o_{i, \pi(y_i)}~,
\end{equation}
where $S_{C}$ denotes the permutation group of $C$ elements. This is essentially taking the the cross entropy loss minimized over all possible permutations of $[C]$.
In our experiments, we consider examples with small numbers of 
communities such as 2 and 5. In general scenarios where $C$ is much larger, the evaluation of the loss function (\ref{lossfunction}) can be impractical due to the minimization over $S_{C}$. A possible solution is to randomly partition $[C]$ into $\tilde{C}$ and then marginalize the model outputs $\{ o_{i,c} \}_{c \in [C]}$ into 
$\{ \tilde{o}_{i, \tilde{c}} := \sum_{c \in \tilde{c}} o_{i,c} \}_{\tilde{c} \in \tilde{C}}$. Finally, we can use 
$\ell( \theta ) =  \min_{ \pi \in S_{\tilde{C}}} - \sum_{i \in V} \log \tilde{o}_{i, \pi(\tilde{y}_i)}$ as an approximate loss value, which only involves a permutation group of size $|\tilde{C}|!$.

Finally, if communities may overlap, 
we can enlarge the label set to include subsets of communities and define the permutation group accordingly. For example, if there are two overlapping communities, we let $\mathcal{C} = \{\{1\}, \{2\}, \{1, 2\}\}$ be the label set, and only allow the permutation between $\{1\}$ and $\{2\}$ when computing the loss function (as well as the overlap to be introduced in Section \ref{experimentalsec}).



\section{Loss landscape of linear GNN optimization}
\label{landscapesec}

As described in the numerical experiments, 
we found that the GNN models without nonlinear activations already provide substantial gains relative to baseline (non-trainable) algorithms.
This section studies the optimization landscape of linear GNNs. Despite defining a non-convex objective, we prove that the landscape is benign under certain further simplifications, in the sense that the local minima are confined in sublevel sets of the loss function. 

For simplicity, we consider only the binary (i.e., $C=2$) case where we replace the node-wise binary cross-entropy loss by the squared cosine distance (which also accounts for the invariance up to a global flip of labels), assume a single dimension of hidden states ($b_k = 1$ for all $k$), and focus on the GNN described in Section \ref{GNNbasic} (although our analysis carries equally to describe the line graph version; see remarks below). We also make the simplifying assumption of replacing the layer-wise instance normalization by a simpler projection onto the unit $\ell_2$ ball (thus we do not remove the mean). 
Without loss of generality, assume that the input graph $G$ has size $n$, and
denote by $\mathcal{F}=\{A_1, \dots, A_Q\}$ the family of graph operators appearing in (\ref{gnneq}). Each layer thus applies an arbitrary polynomial $\sum_{q=1}^Q \theta_q^{(k)} A_q$ to the incoming node feature vector $x^{(k)}$. Given an input node vector $w \in \R^n$, the network output can thus be written as
\begin{equation}
\label{gygy}
\hat{Y} = \frac{e}{\|e\|}\,,\text{ with } e = \left(\prod_{k =1}^K \sum_{q \leq Q} \theta_q^{(k)} A_q \right) w \,.  
\end{equation}
We highlight that this linear GNN setup is fundamentally different from the linear fully-connected neural networks (that is, neural networks with linear activation function), whose landscape has been analyzed in \cite{kawaguchi2016deep}. First, the output of the GNN is on the unit sphere, which has a different geometry. Next, 
the operators in $\mathcal{F}$ depend on the input graph, which introduce fluctuations in the landscape.
In general, the operators in $\mathcal{F}$ are not commutative, but by considering the generalized Krylov subspace generated by powers of $\mathcal{F}$, $\mathcal{F}^K = \{ O_1=A_1^K, O_2=A_1 A_2^{K-1}, O_3=A_1 A_2 A_1^{K-2}, \dots O_{Q^K}=A_Q^K \}$, one can reparametrize (\ref{gygy}) as $e = \sum_{j=1}^{Q^K} \beta_j O_j w$ with $\beta \in \R^{M}$, with $M=Q^K$. Given a graph instance with label vector $y \in \R^n$, the loss it incurs is $1-\frac{| \langle e, y \rangle |^2}{\| e \|^2}$, and therefore the population loss, when expressed as a function of $\beta$, equals
\begin{equation}
\label{loss}
    L_n(\beta) =1-\mathbb{E}_{X_n,Y_n} \frac{\beta^\top Y_n \beta}{ \beta^\top X_n \beta}~,\text{with }
\end{equation}
$$Y_n = z_n z_n^\top \in \R^{M \times M}\,,\,(z_n)_j = \langle O_j w, y \rangle\,\text{ and } X_n = U_n U_n^\top\in \R^{M \times M} \,,U_n = \begin{bmatrix}
(O_1 w)^\top \\
\dots \\
(O_M w)^\top 
\end{bmatrix}\,.
$$
Thus, to study the loss landscape, we examine the properties of the pair of random matrices $Y_n, X_n \in \R^{M \times M}$. Assuming that $\expe X_n \succ 0$, we write the Cholesky decomposition of $\expe X_n$ as $\expe X_n = R_n R_n^T$, and define $A_n = R_n^{-1} Y_n (R_n^{-1})^T$, $\bar{A}_n = \expe A_n = R_n^{-1} \expe Y_n (R_n^{-1})^T$, $B_n = R_n^{-1} X_n (R_n^{-1})^T$, and $\Delta B_n = B_n - I_n$. Given a symmetric matrix $K \in \mathbb{R}^{M \times M}$, we let $\lambda_1(K), \lambda_2(K), ..., \lambda_M(K)$ denote the eigenvalues of $K$ in nondecreasing order. Then, the following theorem establishes that under appropriate assumptions, the concentration of relevant random matrices around their mean controls the energy gaps between local and global minima of $L$.

\begin{theorem} 
\label{main_theorem}
For a given $n$, let $\eta_n = (\lambda_1(\bar{A}_n) - \lambda_2(\bar{A}_n))^{-1}$, $\mu_n = \expe [|\lambda_1(A_n)|^6]$, $\nu_n = \expe [|\lambda_1(B_n)|^{-6}]$, $ \delta_n = \expe [\|  \Delta B_n \|^6]$, and assume that all four quantities are finite.
Then if $\beta_l \in \mathbb{S}^{M-1}$ is a local minimum of $L_n$, and $\beta_g \in \mathbb{S}^{M-1}$ is a global minimum of $L_n$, we have $L_n(\beta_l) \leq (1+\epsilon_{\eta_n, \mu_n, \nu_n, \delta_n}) \cdot L_n(\beta_g)$, where $\epsilon_{\eta_n, \mu_n, \nu_n, \delta_n} = O (\delta_n)$ for given $\eta_n, \mu_n, \nu_n$ as $\delta_n \to 0$ and its formula is given in the appendix.

\end{theorem}

\begin{corollary}
\label{asymptotic}
If $(\eta_n)_{n \in \mathbb{N}^*}$, $(\mu_n)_{n \in \mathbb{N}^*}$, $(\nu_n)_{n \in \mathbb{N}^*}$ are all bounded sequences, and $\lim_{n \to \infty} \delta_n = 0$, then $\forall \epsilon > 0$, $\exists n_\epsilon$ such that $\forall n > n_\epsilon$, $|L_n(\beta_l) - L_n(\beta_g)| \leq \epsilon \cdot L_n(\beta_g)$.
\end{corollary}

The main strategy of the proof is to consider the actual loss function $L_n$ as a perturbation of $\tilde{L}_n(\beta) = 1-\expe_{X_n, Y_n} \frac{\beta^T Y_n \beta}{\beta^T \expe X_n \beta} = 1-\frac{\beta^T \expe Y_n \beta}{\beta^T \expe X_n \beta}$, which has a landscape that is easier to analyze and does not have poor local minima, since it is equivalent to a quadratic form defined over the sphere $\mathbb{S}^{M-1}$. Applying this theorem requires estimating spectral fluctuations of the pair $X_n$, $Y_n$, which in turn involve the spectrum of the $C^*$ algebras generated by the non-commutative family $\mathcal{F}$. For example, for stochastic block models, it is an open problem how the bound behaves as a function of the parameters $p$ and $q$. 
Another interesting question is to understand how the asymptotics of our landscape analysis relate to the hardness of estimation as a function of the signal-to-noise ratio. Finally, another open question is to what extent our result could be extended to the non-linear residual GNN case, perhaps leveraging ideas from \cite{shamir2018resnets}.





\section{Experiments}
\label{experimentalsec}

We present experiments on community detection in synthetic datasets (Sections \ref{SBMsec1}, \ref{SBMsec2} and Appendix \ref{GBMsec}) as well as real-world datasets (Section \ref{snapsec}). In 
the synthetic experiments, the performance is measured by the \textit{overlap} between predicted ($\hat{y}$) and true labels ($y$), which quantifies how much better than random guessing a prediction is, given by 
\begin{equation}
    overlap(y, \hat{y}) = \max_{\pi \in S_C} \big(\frac{1}{n}\sum_u \delta_{\pi(y(u)), \hat{y}(u)} - \frac{1}{C}\big)/(1-\frac{1}{C})
\end{equation}
where $\delta$ is the Kronecker delta, and the maximization is performed over permutations of all the labels. In the real-world datasets, as the communities are overlapping and unbalanced, the prediction accuracy is measured by $\max_{\pi} \frac{1}{n}\sum_u \delta_{\pi(y(u)), \hat{y}(u)}$, and the set of permutations to be maximized over is described in Section \ref{permutationlosssec}.
We use Adamax \citep{adam} with learning rate $0.004$ for optimization across all experiments. The neural network models have 30 layers and 8 dimensions of hidden states in the middle layers (i.e., $b_k = 8$) for experiments in Sections \ref{SBMsec1} and \ref{SBMsec2}, and 20 layers and 6 dimensions of hidden states for Section \ref{snapsec}. GNNs and LGNNs have $J=2$ across the experiments except the ablation experiments in Section \ref{ablations}.  \footnote{The code is available at \url{https://github.com/zhengdao-chen/GNN4CD}}

\subsection{Stochastic Block Models}
\label{SBMsec1}

The stochastic block model (SBM) is a random graph model with planted community structure. A graph sampled from $SBM(n, p, q, C)$ consists of $|V| = n$ nodes partitioned into $C$ communities, that is, each node is assigned a label $y \in \{1, ..., C\}$. An edge connecting any two vertices $u, v$ is drawn independently at random with probability $p$ if $y(v) = y(u)$, and with probability $q$ otherwise. We consider the sparse regime of constant average degree, where $p = a/n$, $q=b/n$ for some $a, b \geq 0$ that do not depend on $n$. As explained in Appendix \ref{app.spectral}, the difficulty of recovering the community labels is indicated by the signal-to-noise ratio (SNR). We compare our GNN and LGNN with belief propagation (BP) as well as spectral methods using the normalized Laplacian and the Bethe Hessian, which we introduce in details in Appendix \ref{sec:background}. In particular, the spectral methods involve performing power iterations for as many times as the number of layers in the GNN and LGNN (which is $30$). We also implement Graph Attention Networks (GAT) as a baseline model\footnote{Implemented based on \url{https://github.com/Diego999/pyGAT}. Similar to our GNN and LGNN, we add instance normalization to every layer. The model contains 30 layers and 8 dimensions of hidden states.}.

\subsubsection{Binary SBM}
For binary SBM (i.e., $C=2$), the SNR has the expression $SNR(a, b) = (a - b)^2 / (2(a+b))$. Thus, we test the models on different choices of SNR by choosing five different pairs of $a_i$ and $b_i$ (or equivalently, $p_i$ and $q_i$) while fixing $a_i + b_i$, thereby maintaining the average degree. In particular, we vary the SNRs around $1$ because for binary SBM under the sparse regime, $SNR = 1$ is the exact threshold for the detection of $y$ to be possible asymptotically in $n$ \citep{abbe2017community}.

\begin{figure}%
\centering
\includegraphics[width=0.48\linewidth,height=2in]{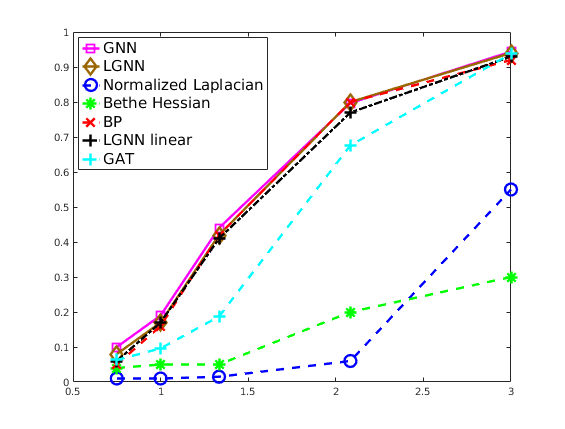}%
\caption{Binary assortative SBM detection (i.e. $C=2$ and $p > q$). X-axis corresponds to SNR, and Y-axis to overlap between the prediction and the ground truth.}
\label{fig:BH1}
\end{figure}

We consider two learning scenarios. In the first scenario, for each pair of $(a_i, b_i)$, we sample $6000$ graphs under $G \sim SBM(n=1000, p_i = a_i / n, q_i = b_i / n, C=2)$ and train the model separately for each $i$.
Figure \ref{fig:BH1} reports the performances of
of the different models in the first learning scenario.
We observe that both GNN and LGNN reach the performance of BP, which is known to be asymptotically optimal \cite{coja2016information} when $p$ and $q$ are known, while GNN and LGNN are agnostic to these parameters.
In addition, even the linear LGNN achieves a performance that is quite close to that of BP, in accordance to the spectral approximations of BP given by the Bethe Hessian (see Appendix \ref{app.spectral}) together with the ability of linear LGNN to express power iterations. These models all significantly outperforms the spectral methods that perform 30 power iterations on the Bethe Hessian or the normalized Laplacian.
We also notice that our models outperform GAT in this task.
%

In the second scenario, whose results are reported in Appendix \ref{sec:furthersbm}, we train a single model from a set of $6000$ graphs sampled from a mixture of SBMs parameterized by the different pairs of $(p_i, q_i)$. This setup demonstrates that our models are more powerful than applying known algorithms such as BP or spectral clustering using Bethe Hessian using SBM parameters learned from data, since the parameters vary in the dataset. 

We also ran experiments in the dissociative case ($q > p$) as well as with $C=3$ communities and obtained similar results, which are not reported here.

\subsection{Probing the computational-to-statistical threshold in $5$-class SBM}
\label{SBMsec2}


In SBM with fewer than 4 communities, it is known 
that BP provably reaches the information-theoretic threshold \citep{abbe2017community,massoulie2014community,coja2016information}. The situation is different for $k>4$, where it is 
conjectured that when the SNR falls into a certain gap, called the \textit{computational-to-statistical gap}, there will be a discrepancy between the theoretical performance of the maximum likelihood estimator and the performance of any polynomial-time algorithm including BP \citep{decelle2011asymptotic}.
In this context, one can use the GNN models to search in the space of generalizations of BP, attempting to improve upon the detection performance of BP for scenarios where the SNR falls within this gap. 
Table \ref{comp-stat-gaps} presents results for the $5$-community disassortative SBM with $n=400$, $p=0$ and $q=18/n$, in which case the SNR lies within the computational-to-statistical gap.
Note that since $p=0$, this also amounts to a graph coloring problem.  

We see that the GNN and LGNN models outperform BP in this experiment, indeed opening up the possibility to bridge the computation-information gap in a data-driven fashion. That said,
our model may be taking advantage of finite-size effects that would vanish as $n\to \infty$. The asymptotic study of these gains is left for future work. In terms of the average test accuracy, LGNN has the best performance. In particular, it outperforms the symmetric version of LGNN, emphasizing the importance of the non-backtracking matrix used in LGNN. Although equipped with the attention mechanism, GAT does not explicitly incorporate in itself the degree matrix, the power graph adjacency matrices or the line graph structure, and has inferior performance compared with the GNN and LGNN models. Further ablation studies on GNN and LGNN are described in Section \ref{ablations}.

\begin{table}[]
\centering
\tabcolsep=0.09cm
\begin{tabular}{l|l|l|l|l|l|l}
\multicolumn{1}{r|}{} & \multicolumn{1}{r|}{GNN} & \multicolumn{1}{r|}{LGNN} & \multicolumn{1}{r|}{LGNN-L} & \multicolumn{1}{r|}{LGNN-S} & \multicolumn{1}{r|}{GAT} & \multicolumn{1}{r}{BP} \\ \hline
Avg. & 0.18 & 0.21 & 0.18 & 0.18 & 0.16 & 0.14 \\
Std. Dev. & 0.04 & 0.05 & 0.04 & 0.04 & 0.04 & 0.02
\end{tabular}
\caption{\footnotesize{Performance of different models on 5-community dissociative SBM graphs with $n=400$, $C=5$, $p=0$, $q=18/n$, corresponding
to an average degree of 14.5}. The first row gives the average overlap across test graphs, and the second row gives the graph-wise standard deviation of the overlap.}
\label{comp-stat-gaps}
\end{table}

\subsection{Real datasets from SNAP}
\label{snapsec}
We now compare the models on the SNAP datasets \citep{snapnets}, whose domains range from social networks to hierarchical co-purchasing networks. 
We obtain the training set as follows. For each SNAP dataset, we select only on the 5000 top quality communities provided in the dataset. We then identify edges $(i,j)$ that cross at least two different communities. For each of such edges, we consider pairs of communities $C_1, C_2$ such that $i \in C_1$, $j \in C_2$, $i \notin C_2$ and $j \notin C_1$, and extract the subset of nodes determined by $C_1 \cup C_2$ together with the edges among them to form a graph. The resulting graph is connected since each community is connected. Finally, we divide the dataset into training and testing sets by enforcing 
that no community belongs to both the training and the testing set.
In our experiment, due to computational limitations, we restrict our attention to the three smallest datasets in the SNAP collection (Youtube, DBLP and Amazon), and we restrict the largest community size to $200$ nodes, which is a conservative bound.

We compare the performance of GNN and LGNN models with GAT as well as the Community-Affiliation Graph Model (AGM), which is a generative model proposed in \cite{JureJaewon_overlapping_com} that captures the overlapping structure of real-world networks.
Community detection can be achieved by fitting AGM to a given network, which was shown to outperform some state-of-the-art algorithms. 
Table \ref{sample-table} compares the performance, measured with a 3-class ($\mathcal{C} = \{ \{1\},\{2\}, \{1, 2\}\}$) classification accuracy up to global permutation of $\{1\} \leftrightarrow \{2\}$. GNN, LGNN, LGNN-S and GAT yield similar results and outperform AGMfit, with the first three achieving the highest average accuracies. It further illustrates the benefits of data-driven models that strike the right balance 
between expressivity and structural design.

\begin{table}[h]
\centering
{\small 

\begin{tabular}{l|l|l|l|l|l|l|l|l|l}
        & train/test & Avg |V| & Avg |E| &           & GNN  & LGNN & LGNN-S & GAT  & AGMfit \\ \hline
Amazon  & 805/142    & 60       & 161      & Avg.      & 0.97 & 0.96 & 0.97     & 0.95 & 0.90   \\ \cline{5-10} 
        &            &          &          & Std. Dev. & 0.12 & 0.13 & 0.11     & 0.13 & 0.13   \\ \hline
DBLP    & 4163/675   & 26       & 77       & Avg.      & 0.90 & 0.90 & 0.89     & 0.88 & 0.79   \\ \cline{5-10} 
        &            &          &          & Std. Dev. & 0.13 & 0.13 & 0.13     & 0.13 & 0.18   \\ \hline
Youtube & 20000/1242 & 93       & 201      & Avg.      & 0.91 & 0.92 & 0.91     & 0.90 &  0.59      \\ \cline{5-10} 
        &            &          &          & Std. Dev. & 0.11 & 0.11 & 0.11     & 0.13 &  0.16     
\end{tabular}
  }
  \caption{\footnotesize{Comparison of the node classification accuracy by different models on the three SNAP datasets. Note that the average accuracy is computed graph-wise with each graph weighted by its size, while the standard deviation is computed graph-wise with equal weights among the graphs.}}
  \label{sample-table}
  \centering
\end{table}
\section{Conclusions}
In this work, we have studied data-driven approaches to supervised community detection with graph neural networks. 
Our models achieve similar performance to BP in binary SBM for various choices of SNR in the data, and outperform BP in the sparse regime of 5-class SBM that falls between the computational-to-statistical gap. This is made possible by considering a family of graph operators including the power graph adjacency matrices, and importantly by introducing the line graph equipped with the non-backtracking matrix. We also provide a theoretical analysis of the optimization landscapes of simplified linear GNN for community detection and showed the gap between the loss values at local and global minima are bounded by quantities related to the concentration of certain random matricies. 

One word of caution is that our empirical results are inherently non-asymptotic. Whereas models trained for given graph sizes can be used for inference on arbitrarily sized graphs (owing to the parameter sharing of GNNs), further work is needed in order to understand the generalization properties as $|V|$ increases. 
Nevertheless, we believe our work the study of computational-to-statistical gaps, where our model can be used to inquire about the form of computationally tractable approximations. Moreover, our work also opens up interesting questions, including whether the network parameters can be interpreted mathematically, and how our results on the energy landscape depend upon specific signal-to-noise ratios. 
Another current limitation of our model is that it presumes a fixed number of communities to be detected. Thus, other directions of future research include the extension to the case where the number of communities is unknown and varied, or even increasing with $|V|$, as well as applications to ranking and edge-cut problems. 


\section{Acknowledgement}
This work was partially supported by the Alfred P. Sloan Foundation and DOA W911NF-17-1-0438.

\bibliography{nips}

\bibliographystyle{iclr2019_conference}

\appendix

\section{Proof of Theorem 5.1}




For simplicity and with an abuse of notation, in the remaining part we redefine $L$ and $\tilde{L}$ in the following way, to be the negative of their original definition in the main section: $L_n(\beta) =\mathbb{E}_{X_n,Y_n} \frac{\beta^\top Y_n \beta}{ \beta^\top X_n \beta}$, $\tilde{L}_n(\beta) = \expe_{X_n, Y_n} \frac{\beta^T Y_n \beta}{\beta^T \expe X_n \beta}$. Thus, minimizing the loss function (\ref{loss}) is equivalent to maximizing the function $L_n(\beta)$ redefined here.

We write the Cholesky decomposition of $\expe X_n$ as $\expe X_n = R_n R_n^T$, and define $A_n = R_n^{-1} Y_n (R_n^{-1})^T$, $\bar{A}_n = \expe A_n = R_n^{-1} \expe Y_n (R_n^{-1})^T$, $B_n = R_n^{-1} X_n (R_n^{-1})^T$, and $\Delta B_n = B_n - I_n$. Given a symmetric matrix $K \in \mathbb{R}^{M \times M}$, we let $\lambda_1(K), \lambda_2(K), ..., \lambda_M(K)$ denote the eigenvalues of $K$ in nondecreasing order.

First, we have
\begin{equation}
\label{first_bound}
    |L_n(\beta_l) - L_n(\beta_g)| \leq |L_n(\beta_l) - \tilde{L}_n(\beta_l)| + |\tilde{L}_n(\beta_l) - \tilde{L}_n(\beta_g)| + |\tilde{L}_n(\beta_g) - L_n(\beta_g)|
\end{equation}
Let us denote by $\tilde{\beta_g}$ a global minimum 
of the mean-field loss $\tilde{L}_n$. 
Taking a step further, we can extend this bound to the following one (the difference is in the second term on the right hand side):

\begin{lemma}
\label{break_into_three}
\begin{equation}
    |L_n(\beta_l) - L_n(\beta_g)| \leq |L_n(\beta_l) - \tilde{L}_n(\beta_l)| + |\tilde{L}_n(\beta_l) - \tilde{L}_n(\tilde{\beta}_g)| + |\tilde{L}_n(\beta_g) - L_n(\beta_g)|
\end{equation}
\end{lemma}

\begin{proof}[Proof of Lemma \ref{break_into_three}]
We consider two separate cases: The first case is when $\tilde{L}_n(\beta_l) \geq \tilde{L}_n(\beta_g)$. Then $\tilde{L}_n(\beta_l) - \tilde{L}_n(\tilde{\beta}_g) \geq \tilde{L}_n(\beta_l) - \tilde{L}_n(\beta_g) \geq 0$, and so 
$
|L_n(\beta_l) - L_n(\beta_g)| \leq |L_n(\beta_l) - \tilde{L}_n(\beta_l)| + |\tilde{L}_n(\beta_l) - \tilde{L}_n(\tilde{\beta}_g)| + |\tilde{L}_n(\beta_g) - L_n(\beta_g)|
$.

The other case is when $\tilde{L}_n(\beta_l) < \tilde{L}_n(\beta_g)$. Note that $L_n(\beta_l) \geq L_n(\beta_g)$. Then $|L_n(\beta_l) - L_n(\beta_g)| \leq |L_n(\beta_l) - \tilde{L}_n(\beta_l)| + |\tilde{L}_n(\beta_g) - L_n(\beta_g)| \leq |L_n(\beta_l) - \tilde{L}_n(\beta_l)| + |\tilde{L}_n(\beta_l) - \tilde{L}_n(\tilde{\beta}_g)| + |\tilde{L}_n(\beta_g) - L_n(\beta_g)|$.

\end{proof}

Hence, to bound the "energy gap" $|L_n(\beta_l) - L_n(\beta_g)|$, if suffices to bound the three terms on the right hand side of Lemma \ref{break_into_three} separately. First, we consider the second term, $|\tilde{L}_n(\beta_l) - \tilde{L}_n(\tilde{\beta}_g)|$.

Let $\gamma_l = R_n^T \beta_l, \gamma_g = R_n^T \beta_g$ and $\tilde{\gamma}_g = R_n^T \tilde{\beta}_g$. Define $S_n(\gamma) = L_n(R_n^{-T} \gamma)$ and $\tilde{S}_n(\gamma) = \tilde{L}_n(R_n^{-T} \gamma)$, for any $\gamma \in \mathbb{R}^M$. Thus, we apply a change-of-variable and try to bound $|\tilde{S}_n(\gamma_l) - \tilde{S}_n(\tilde{\gamma}_g)|$.

Since $\beta_l$ is a local maximum of $L_n$, $\lambda_{1}(\nabla^2 L_n(\beta_l)) \leq 0$. Since $\nabla^2 S_n(\gamma_l) = R_n^{-1} \nabla^2 L_n(\beta_l) R_n^{-T}$, where $R_n$ is invertible, we know that $\lambda_{1}(\nabla^2 S_n(\gamma_l)) \leq 0$, thanks to the following lemma:

\begin{lemma}
\label{change_of_var}
If $R, Q \in \mathbb{R}^{M \times M}$, $R$ is invertible, $Q$ is symmetric and $\lambda_q > 0$ is an eigenvalue of $Q$, then $\lambda_{1} (R Q R^T) \geq \lambda \cdot \lambda_{M}(RR^T)$
\end{lemma}

\begin{proof}[Proof of Lemma \ref{change_of_var}]
Say $Q w = \lambda w$ for some vector $w \in \mathbb{R}^M$. Let $v = R^{-T} w$. Then $v^T (R Q R^T) v = w^T Q w = \lambda \| w \|^2$. Note that 
$\| w \|^2 = v^T R R^T v \geq \| v \|^2 \lambda_M(R R^T)$.
Hence
$\lambda_{1} (R Q R^T) \geq \frac{v^T (R Q R^T) v}{\| v \|^2} \geq \frac{\lambda \| w \|^2}{\| w \|^2 / \lambda_M(R R^T)} \geq \lambda \cdot \lambda_{M}(RR^T)$
\end{proof}

Since \(\nabla^2 S_n(\gamma_l) = \nabla^2 \tilde{S}_n(\gamma_l) + (\nabla^2 S_n(\gamma_l) - \nabla^2 \tilde{S}_n(\gamma_l))\), there is \(0 \geq \lambda_{1}(\nabla^2 S_n(\gamma_l)) \geq \lambda_{1}(\nabla^2 \tilde{S}_n(\gamma_l)) - \|\nabla^2 S_n(\gamma_l) - \nabla^2 \tilde{S}_n(\gamma_l)\|\).
Hence,
\begin{equation}
\label{first_ineq}
\lambda_{1}(\nabla^2 \tilde{S}_n(\gamma_l)) \leq \|\nabla^2 S_n(\gamma_l) - \nabla^2 \tilde{S}_n(\gamma_l)\|
\end{equation}

    

Next, we relate the left hand side of the inequality above to $\cos(\gamma_l, \tilde{\gamma}_g)$, thereby obtaining an upper bound on $[1 - \cos^2(\gamma_l, \tilde{\gamma}_g)]$, which will then be used to bound $|\tilde{S}_n(\gamma_l) - \tilde{S}_n(\tilde{\gamma}_g)|$.

\begin{lemma} 
\label{bound_noiseless_hessian}
$\forall \gamma \in \mathbb{R}^d$,
\[\lambda_{1}(\nabla^2 \tilde{S}_n(\gamma)) \geq \frac{2}{\| \gamma \|^2}\{[1 - \cos^2(\gamma, \tilde{\gamma}_g)] \cdot [\lambda_1(\bar{A}_n) - \lambda_2(\bar{A}_n)] - 2 \| \gamma\| \cdot \| \nabla \tilde{S}_n(\gamma) \|\}
\]
\end{lemma}

\begin{proof}[Proof of Lemma \ref{bound_noiseless_hessian}]
\begin{equation}
\label{noiseless_hess}
\begin{split}
    \nabla^2 \tilde{S}_n(\gamma) =& 2 \expe \left[\frac{(\gamma^T \gamma) A_n - (\gamma^T A_n \gamma) I}{(\gamma^T \gamma)^2} + \frac{4 (\gamma^T A_n \gamma) \gamma \gamma^T - 4 (\gamma^T \gamma) A_n \gamma \gamma^T}{(\gamma^T \gamma)^3}\right]\\
    =& 2 \expe \left[\frac{(\gamma^T \gamma) A_n - (\gamma^T A_n \gamma) I}{(\gamma^T \gamma)^2} + \frac{4 [(\gamma^T \gamma) A_n - (\gamma^T A_n \gamma) I] \gamma \gamma^T }{(\gamma^T \gamma)^3}\right] \\
    =& 2 \left[\frac{(\gamma^T \gamma) \bar{A}_n - (\gamma^T \bar{A}_n \gamma) I}{(\gamma^T \gamma)^2} + \frac{4 [(\gamma^T \gamma) \bar{A}_n - (\gamma^T \bar{A}_n \gamma) I] \gamma \gamma^T }{(\gamma^T \gamma)^3}\right]
\end{split}
\end{equation}
Thus, if we define $Q_1 = (\gamma^T \gamma)[(\gamma^T \gamma) \bar{A}_n - (\gamma^T \bar{A}_n \gamma) I]$, $Q_2 = 4 [(\gamma^T \gamma) \bar{A}_n - (\gamma^T \bar{A}_n \gamma) I] \gamma \gamma^T$, we have 
\begin{equation}
    \nabla^2 \tilde{S}_n(\gamma) = \frac{2}{\| \gamma \|^6} (Q_1 - Q_2)
\end{equation}
To bound $\lambda_1(\nabla^2 \tilde{S}_n(\gamma))$, we bound $\lambda_1 (Q_1)$ and $\| Q_2\|$ as follows:

Since $\bar{A}_n$ is symmetric, let $\hat{\gamma}_{1}, \dots \hat{\gamma}_{M}$ be the orthonormal eigenvectors of $\bar{A}_n$ corresponding to nonincreasing eigenvalues $l_1, \dots l_M$. Note that the global minimum satisfies $\tilde{\gamma}_g = \pm \hat{\gamma}_1$. Write $\gamma = \sum_{i=1}^M \alpha_i \hat{\gamma}_i$, and let $\bar{\alpha}_i = \frac{\alpha_i}{\sqrt{\sum_{i=1}^M \alpha_i^2}}$. Then $|\cos(\gamma, \tilde{\gamma}_g)| = |\cos(\gamma, \hat{\gamma}_1)| = |\bar{\alpha}_1|$.

Then,
\begin{equation}
    \begin{split}
        \lambda_{1}(Q_1) =& (\gamma^T \gamma) \left[l_1 \sum_{i=1}^M \alpha_i^2 - \sum_{i=1}^M l_i \alpha_i^2\right] \\
        \geq& (\gamma^T \gamma)\left[\left((\sum_{i=1}^M \alpha_i^2) - \alpha_1^2\right) (l_1 - l_2)\right]\\
        =& (\gamma^T \gamma)^2 [(1 - \bar{\alpha}_1^2) (l_1 - l_2)]
    \end{split}
\end{equation}

To bound $\|Q_2\|$:
\begin{equation}
    \begin{split}
        [(\gamma^T \gamma) \bar{A}_n - (\gamma^T \bar{A}_n \gamma) I] \gamma 
        = \sum_{k=1}^M \left[l_k \sum_{i=1}^M \alpha_i^2 - \sum_{i=1}^M l_i \alpha_i^2\right] \alpha_k \hat{\gamma}_k 
    \end{split}
\end{equation}
Note that given vectors $v, w \in \mathbb{R}^M$, 
\[
\| v \cdot w^T \| = |v^T w|
\]

Therefore, 
\begin{equation}
    \begin{split}
        \|Q_2\| =& 4 \left|\left(\sum_{k=1}^M \alpha_k \hat{\gamma}_k\right)^T \left(\sum_{k=1}^M [l_k(\sum_{i=1}^M \alpha_i^2) - (\sum_{i=1}^M l_i \alpha_i^2)] \alpha_k \hat{\gamma}_k \right)\right| \\
        =& 4 \left|\frac{(\gamma^T \gamma)^2}{2} \gamma^T \nabla \tilde{S}(\gamma)\right| \\
        \leq & 2 (\gamma^T \gamma)^2 \| \gamma \| \| \nabla \tilde{S}(\gamma)\|
    \end{split}
\end{equation}

Thus, 
\begin{equation}
    \begin{split}
        \lambda_{1}(Q_1 - Q_2) \geq& \lambda_{1}(Q_1) - \| Q_2\| \\
        \geq& (\gamma^T \gamma)^2 ([(1 - \bar{\alpha}_1^2) (l_1 - l_2)] - 2 \| \gamma \| \| \nabla_\gamma S(\gamma)\|) 
    \end{split}
\end{equation}
This yields the desired lemma.
\end{proof}

Combining inequality \ref{first_ineq} and Lemma \ref{bound_noiseless_hessian}, we get
\begin{equation}
\label{bound_on_cos}
1-\cos^2(\gamma_l, \tilde{\gamma}_g) \leq \frac{ 2 \| \gamma_l \| \cdot \| \nabla \tilde{S}_n(\gamma_l)\| + \frac{\| \gamma_l \|^2}{2} \|\nabla^2 S_n(\gamma_l) - \nabla^2 \tilde{S}_n(\gamma_l)\|}{\lambda_1(\bar{A}_n)- \lambda_2(\bar{A}_n)}
\end{equation}

Thus, to bound the angle between $\gamma_l$ and $\tilde{\gamma}_g$, we can aim to bound $\| \nabla \tilde{S}_n(\gamma_l)\|$ and $\|\nabla^2 S_n(\gamma_l) - \nabla^2 \tilde{S}_n(\gamma_l)\|$ as functions of the quantities $\mu_n$, $\nu_n$ and $\delta_n$.




\begin{lemma}
\label{perturbation_gradient_nonasymptotic}
\begin{equation}
    \|\gamma_l \| \cdot \| \nabla \tilde{S}_n(\gamma_l)\| \leq 2 \mu_n \nu_n \delta_n (1 + 3 \nu_n + \delta \nu_n)
\end{equation}
\end{lemma}

\begin{proof}[Proof of Lemma \ref{perturbation_gradient_nonasymptotic}]
\begin{equation}
\label{noised_grad}
    \nabla S_n(\gamma) = 2 \expe \frac{A_n \gamma}{\gamma^T B_n \gamma} - 2 \expe \frac{(\gamma^T A_n \gamma) B_n \gamma}{(\gamma^T B_n \gamma)^2}
\end{equation}
\begin{equation}
\label{noiseless_grad}
    \nabla \tilde{S}_n(\gamma) = 2 \expe \frac{A_n \gamma}{\gamma^T \gamma} - 2 \expe \frac{(\gamma^T A_n \gamma) \gamma}{(\gamma^T \gamma)^2}
\end{equation}
Combining equations \ref{noised_grad} and \ref{noiseless_grad}, we get 
\begin{equation}
    \nabla S_n(\gamma) - \nabla \tilde{S}_n(\gamma) =
    \expe \left[ \frac{2 (\gamma^T \gamma - \gamma^T B_n \gamma) A_n \gamma}{(\gamma^T B_n \gamma)(\gamma^T \gamma)} - \frac{2 (\gamma^T A_n \gamma)[(\gamma^T \gamma)^2 B_n \gamma - (\gamma^T B_n \gamma)^2 \gamma]}{(\gamma^T B_n \gamma)^2 (\gamma^T \gamma)^2} \right]
\end{equation}
Since $\nabla S_n(\gamma_l) = 0$, we have 
\begin{equation}
    \begin{split}
        \| \nabla \tilde{S}_n(\gamma_l) \| =& \left\| \expe \left[ \frac{2 (\gamma_l^T \gamma_l - \gamma_l^T B_n \gamma_l) A_n \gamma_l}{(\gamma_l^T B_n \gamma_l)(\gamma_l^T \gamma_l)} - \frac{2 (\gamma_l^T A_n \gamma_l)[(\gamma_l^T \gamma_l)^2 B_n \gamma_l - (\gamma_l^T B_n \gamma_l)^2 \gamma_l]}{(\gamma_l^T B_n \gamma_l)^2 (\gamma_l^T \gamma_l)^2} \right] \right\| \\
        \leq& \frac{2}{\| \gamma_l \|} \expe \left[\frac{|\lambda_1(A_n)| \| \Delta B_n\|}{|\lambda_M(B_n)|} + 3 \frac{|\lambda_1(A_n)| \| \Delta B_n\|}{\lambda_M^2(B_n)} + \frac{|\lambda_1(A_n)| \| \Delta B_n\|^2}{\lambda_M^2(B_n)} \right]
    \end{split}
\end{equation}
Then, by the generalized H{\"o}lder's inequality,

\begin{equation}
\begin{split}
\| \nabla \tilde{S}_n(\gamma_l) \| \leq & \frac{2}{\| \gamma_l \|} \Big[\left(\expe |\lambda_1(A_n)|^3 \expe \|\Delta B_n \|^3 \expe  \frac{1}{|\lambda_M(B_n)|^3}\right)^{\frac{1}{3}} \\
&+ 3 \left(\expe |\lambda_1(A_n)|^3 \expe \|\Delta B_n \|^3 \expe  \frac{1}{|\lambda_M(B_n)|^6}\right)^{\frac{1}{3}} \\
 &+ \left(\expe |\lambda_1(A_n)|^3 \expe \|\Delta B_n \|^6 \expe  \frac{1}{|\lambda_M(B_n)|^6}\right)^{\frac{1}{3}} \Big]~.
\end{split}
\end{equation}

Hence, written in terms of the quantities $\mu_n$, $\nu_n$ and $\delta_n$, we have

\begin{equation}
\begin{split}
\| \gamma_l \| \cdot \| \nabla \tilde{S}_n(\gamma_l) \| \leq & 
2 (\mu_n \nu_n \delta_n + 3 \mu_n \nu_n^2 \delta_n + \mu_n \delta_n^2 \nu_n^2) \\
=& 2 \mu_n \nu_n \delta_n (1 + 3 \nu_n + \delta \nu_n)
\end{split}
\end{equation}


\end{proof}

\begin{lemma}
\label{delta_1}
With $\delta_n = (\expe \| \Delta B_n \|^6)^\frac{1}{6}$, $\expe |\lambda_1(B_n)|^6 \leq 64 + 63 \delta_n^6$
\end{lemma}
\begin{proof}[Proof of Lemma \ref{delta_1}]
\begin{equation}
    \begin{split}
        \expe |\lambda_1(B_n)|^6 = & \expe \| B_n \|^6 \\
        = & \expe \| I + \Delta B_n \|^6 \\
        \leq & \expe (\| I \| + \| \Delta B_n \|)^6 \\
        = & \expe (1 + \| \Delta B_n \|)^6
    \end{split}
\end{equation}
Note that
\begin{equation}gma
    \begin{split}
        \expe (1 + X)^6 = \expe X^6 + 6 \expe X^5 + 15 \expe X^4 + 20 \expe X^3 + 15 \expe X^2 + 6 \expe X + 1 
    \end{split}
\end{equation}
and for $k \in \{1, 2, 3, 4, 5\}$, if $X$ is a nonnegative random variable,
\begin{equation}
    \begin{split}
        \expe X^k = & \mathds{1}_{X > 1} \expe X^k + \mathds{1}_{X \leq 1} \expe X^k \\
        \leq & 1 + \mathds{1}_{X \leq 1} \expe X^6 \\
        \leq & 1 + \expe X^6
    \end{split}
\end{equation}
Therefore, $\expe |\lambda_1(B_n)|^6 \leq 64 + 63 \expe \| \Delta B_n \|^6$.
\end{proof}

From now on, for simplicity, we introduce $\delta'_n = (64 + 63 \delta_n^6)^\frac{1}{6}$, as a function of $\delta_n$.

\begin{lemma}
\label{perturbation_hessian_nonasymptotic}
$\forall \gamma \in \mathbb{R}^{M}$,
\begin{equation}
\begin{split}
\| \gamma_l \|^2 \cdot \|\nabla^2 S_n(\gamma) - \nabla^2 \tilde{S}_n(\gamma) \| \leq & \mu_n \nu_n \delta_n (10 + 14 \nu_n + 2 \delta_n \nu_n + 16 \nu_n^2 + 16 \delta'_n \nu_n \\ & + 8 \delta'_n \nu_n^2 + 8 \delta'_n \nu_n + 8 \delta_n \delta'_n \nu)
\end{split}
\end{equation}
\end{lemma}

\begin{proof}[Proof of Lemma \ref{perturbation_hessian_nonasymptotic}]
\begin{equation}
    \begin{split}
        \nabla^2 S_n(\gamma) - \nabla^2 \tilde{S}_n(\gamma) =&
        2 \expe [H_1] - 2 \expe [H_2] + 8 \expe [H_2] - 8 \expe [H_4]
    \end{split}
\end{equation}
where
\begin{equation}
    H_1 = \frac{(\gamma^T \gamma) A_n - (\gamma^T B_n \gamma) A_n}{(\gamma^T B_n \gamma)(\gamma^T \gamma)}
\end{equation}
\begin{equation}
    H_2 = \frac{(\gamma^T A_n \gamma) [(\gamma^T \gamma)^2 B_n - (\gamma^T B_n \gamma)^2] I)}{(\gamma^T B \gamma)^2 (\gamma^T \gamma)^2}
\end{equation}
\begin{equation}
    H_3 = \frac{(\gamma^T A_n \gamma) [(\gamma^T \gamma)^3 B_n \gamma \gamma^T B_n^T - (\gamma^T B_n \gamma)^3 \gamma \gamma^T]}{(\gamma^T B_n \gamma)^3 (\gamma^T \gamma)^3}
\end{equation}
\begin{equation}
    H_4 = \frac{(\gamma^T \gamma)^2 A_n \gamma \gamma^T B_n - (\gamma^T B_n \gamma)^2 A \gamma \gamma^T}{(\gamma^T B_n \gamma)^2 (\gamma^T \gamma)^2}
\end{equation}
Thus, $\|\nabla^2 S_n(\gamma) - \nabla^2 \tilde{S}_n(\gamma) \| \leq 2 \expe \|H_1\| + 2 \expe \|H_2\| + 8 \expe \|H_3\| + 8 \expe \|H_4\|$, and we try to bound each term on the right hand side separately.

For the first term, there is
\begin{equation}
    \| H_1 \| \leq \frac{1}{\| \gamma \|^2} \frac{\| \Delta B_n \| |\lambda_1(A_n)|}{|\lambda_M(B_n)|}
\end{equation}

Applying generalized H{\"o}lder's inequality, we obtain
\begin{equation}
    \begin{split}
        \|\gamma\|^2 \cdot \expe \| H_1\| \leq &  \left(\expe \frac{1}{|\lambda_M(B_n)|^3}\right)^\frac{1}{3} (\expe |\lambda_1(A_n)|^3)^\frac{1}{3} (\expe \| \Delta B_n \|^3)^\frac{1}{3} \\
        \leq & \mu_n \nu_n \delta_n~.
    \end{split}
\end{equation}

For the second term, there is 
\begin{equation}
    H_2 = \frac{(\gamma^T A_n \gamma)[(\gamma^T \gamma)^2 \Delta B_n - 2 (\gamma^T \gamma) (\gamma^T \Delta B_n \gamma) I - (\gamma^T \Delta B_n \gamma)^2 I]}{(\gamma^T B_n \gamma)^2 (\gamma^T \gamma)^2}
\end{equation}

Hence, 
\begin{equation}
    \| H_2 \| \leq \frac{1}{\| \gamma \|^2} \frac{1}{\lambda_M^2 (B_n)} |\lambda_1(A_n)| (3 \| \Delta B_n \| + \| \Delta B_n \|^2)
\end{equation}

Applying generalized H{\"o}lder's inequality, we obtain
\begin{equation}
    \begin{split}
        \|\gamma\|^2 \cdot \expe \| H_2\| \leq &  \left(\expe \frac{3}{|\lambda_M(B_n)|^6}\right)^\frac{1}{3} (\expe |\lambda_1(A_n)|^3)^\frac{1}{3} (\expe \| \Delta B_n \|^3)^\frac{1}{3} \\
        & + \left(\expe \frac{3}{|\lambda_M(B_n)|^6}\right)^\frac{1}{3} (\expe |\lambda_1(A_n)|^3)^\frac{1}{3} (\expe \| \Delta B_n \|^6)^\frac{1}{3} \\
        \leq & \mu_n \nu_n \delta_n (3 \nu_n + \delta_n \nu_n)
    \end{split}
\end{equation}

For $H_3$, note that
\begin{equation}
\begin{split}
    (\gamma^T \gamma)^3 B_n \gamma \gamma^T B_n^T - (\gamma^T B_n \gamma)^3 \gamma \gamma^T =& 
    (\gamma^T \gamma)^3 (B_n - I) \gamma \gamma^T B_n + (\gamma^T \gamma)^3 \gamma \gamma^T (B_n - I)\\
    & + [(\gamma^T \gamma)^3 - (\gamma^T B_n \gamma)^3] \gamma \gamma^T \\
    =& (\gamma^T \gamma)^3 \Delta B_n \gamma \gamma^T B_n + (\gamma^T \gamma)^3 \gamma \gamma^T \Delta B_n \\
    & + [(\gamma^T B_n \gamma)^2 (-\gamma^T \Delta B_n \gamma) \gamma \gamma^T + (\gamma^T B_n \gamma) (-\gamma^T \Delta B_n \gamma) \gamma \gamma^T \\
    & + (-\gamma^T \Delta B_n \gamma) \gamma \gamma^T]
\end{split}
\end{equation}
Hence, 
\begin{equation}
    \begin{split}
        H_3 =& (\gamma^T A_n \gamma) \Big[\frac{(\gamma^T \gamma)^3 \Delta B_n \gamma \gamma^T B_n + (\gamma^T \gamma)^3 \gamma \gamma^T \Delta B_n + (-\gamma^T \Delta B_n \gamma) \gamma \gamma^T}{(\gamma^T B_n \gamma)^3 (\gamma^T \gamma)^3} \\
        & + \frac{(-\gamma^T \Delta B_n \gamma) \gamma \gamma^T}{(\gamma^T B_n \gamma)^2(\gamma^T \gamma)} + \frac{(-\gamma^T \Delta B_n \gamma) \gamma \gamma^T}{(\gamma^T B_n \gamma)(\gamma^T \gamma)^2}\Big]
    \end{split}
\end{equation}
Thus, 
\begin{equation}
    \|H_3\| \leq \frac{|\lambda_1(A_n)|}{\| \gamma \|^2} \left[\frac{1}{|\lambda_M^3(B_n)|}(\| \Delta B_n \| |\lambda_1(B_n)| + 2 \| \Delta B_n \|) + \frac{1}{\lambda_M^2(B_n)} \| \Delta B_n\| + \frac{1}{|\lambda_M(B_n)|}\| \Delta B_n \|\right]
\end{equation}

Applying generalized H{\"o}lder's inequality, we obtain
\begin{equation}
    \begin{split}
        \|\gamma\|^2 \cdot \expe \| H_3\| \leq & \left(\expe \frac{1}{|\lambda_M(B_n)|^6}\right)^\frac{1}{2} (\expe |\lambda_1(A_n)|^6)^\frac{1}{6} (\expe \| \Delta B_n \|^6)^\frac{1}{6} (\expe |\lambda_1(B_n)|^6)^\frac{1}{6} \\
        &+ 2\left(\expe \frac{1}{|\lambda_M(B_n)|^6}\right)^\frac{1}{2} (\expe |\lambda_1(A_n)|^3)^\frac{1}{3} (\expe \| \Delta B_n \|^6)^\frac{1}{6} \\
        &+ \left(\expe \frac{1}{|\lambda_M(B_n)|^6}\right)^\frac{1}{3} (\expe |\lambda_1(A_n)|^3)^\frac{1}{3} (\expe \| \Delta B_n \|^3)^\frac{1}{3} \\
        &+ \left(\expe \frac{1}{|\lambda_M(B_n)|^3}\right)^\frac{1}{3} (\expe |\lambda_1(A_n)|^3)^\frac{1}{3} (\expe \| \Delta B_n \|^3)^\frac{1}{3} \\
        \leq & \mu_n \nu_n \delta_n (\delta'_n \nu_n^2 + 2 \nu_n^2 + \nu_n + 1)
    \end{split}
\end{equation}

For the last term,

\begin{equation}
    H_4 = \frac{[-2(\gamma^T \gamma)(\gamma^T \Delta B_n \gamma) I - (\gamma^T \Delta B_n \gamma)^2 I] A_n \gamma \gamma^T B_n + (\gamma^T B_n \gamma)^2 A_n \gamma \gamma^T \Delta B_n}{(\gamma^T B_n \gamma)^2 (\gamma^T \gamma)^2}
\end{equation}

Thus, \begin{equation}
    \| H_4 \| \leq \frac{1}{\| \gamma \|^2} \left[ \frac{1}{\lambda_M^2(B_n)}(2 \| \Delta B_n \| + \| \Delta B_n \|^2) |\lambda_1(A_n)| |\lambda_1(B_n)| + \frac{1}{\lambda_M^2(B_n)} |\lambda_1^2 (B_n)| |\lambda_1(A_n)| \| \Delta B_n\| \right]
\end{equation}

Applying generalized H{\"o}lder's inequality, we obtain
\begin{equation}
\begin{split}
    \|\gamma\|^2 \cdot \expe \| H_4 \| \leq & 2 \left(\expe \frac{1}{|\lambda_M(B_n)|^6}\right)^\frac{1}{3} (\expe |\lambda_1(A_n)|^3)^\frac{1}{3} (\expe \| \Delta B_n \|^6)^\frac{1}{6} (\expe |\lambda_1(B_n)|^6)^\frac{1}{6} \\
    & + \left(\expe \frac{1}{|\lambda_M(B_n)|^6}\right)^\frac{1}{3} (\expe |\lambda_1(A_n)|^6)^\frac{1}{6} (\expe \| \Delta B_n \|^6)^\frac{1}{3} (\expe |\lambda_1(B_n)|^6)^\frac{1}{6} \\
    & +  \left(\expe \frac{1}{|\lambda_M(B_n)|^6}\right)^\frac{1}{3} (\expe |\lambda_1(A_n)|^6)^\frac{1}{6} (\expe \| \Delta B_n \|^6)^\frac{1}{6} (\expe |\lambda_1(B_n)|^6)^\frac{1}{3} \\
    \leq & \mu_n \nu_n \delta_n (2 \nu_n \delta'_n + \delta_n \delta'_n \nu_n + {\delta'_n}^2 \nu_n)
\end{split}
\end{equation}

Therefore, summing up the bounds above, we obtain
\begin{equation}
    \begin{split}
        \| \gamma_l \|^2 \cdot \|\nabla^2 S_n(\gamma) - \nabla^2 \tilde{S}_n(\gamma) \|
        \leq & \mu_n \nu_n \delta_n (10 + 14 \nu_n + 2 \delta_n \nu_n + 16 \nu_n^2 + 16 \delta'_n \nu_n \\ & + 8 \delta'_n \nu_n^2 + 8 \delta'_n \nu_n + 8 \delta_n \delta'_n \nu)
    \end{split}
\end{equation}

Hence, combining inequality \ref{bound_on_cos}, Lemma \ref{perturbation_gradient_nonasymptotic} and Lemma \ref{perturbation_hessian_nonasymptotic}, we get
\begin{equation}
    \begin{split}
        1-\cos^2(\gamma_l, \tilde{\gamma}_g) \leq & \eta_n [4 \mu_n \nu_n \delta_n (1 + 3 \nu_n \delta_n \mu_n) + \frac{1}{2}\mu_n \nu_n \delta_n (10 + 14 \nu_n + 2 \delta_n \nu_n + 16 \nu_n^2 \\ & + 16 \delta'_n \nu_n + 8 \delta'_n \nu_n^2 + 8 \delta'_n \nu_n + 8 \delta_n \delta'_n \nu)] \\
        = & \mu_n \nu_n \delta_n \eta_n ( 9 + 19 \nu_n + 5 \delta_n \nu_n + 8 \nu_n^2 + 8 \delta'_n \nu_n + 4 \delta'_n {\nu_n}^2 + 4 \delta'_n \nu_n + 4 \delta_n \delta'_n \nu_n)
    \end{split}
\end{equation}
For simplicity, we define $C(\delta_n, \nu_n) = 9 + 19 \nu_n + 5 \delta_n \nu_n + 8 \nu_n^2 + 8 \delta'_n \nu_n + 4 \delta'_n {\nu_n}^2 + 4 \delta'_n \nu_n + 4 \delta_n \delta'_n \nu_n$. \\
Thus, 
\begin{equation}
\label{bound_cos_2}
    1-\cos^2(\gamma_l, \tilde{\gamma}_g) \leq \mu_n \nu_n \delta_n \eta_n C(\delta_n, \nu_n)
\end{equation}


\end{proof}

Following the notations in the proof of Lemma \ref{bound_noiseless_hessian}, we write $\gamma_l = \sum_{i=1}^M \alpha_i \hat{\gamma}_i$. Note that $\tilde{\gamma}_g = \pm \hat{\gamma}_1$, and $|\cos(\gamma, \hat{\gamma}_i)| = |\bar{\alpha}_i|$. Thus, 
\begin{equation}
    \begin{split}
        \tilde{L}_n(\beta_l) =& \tilde{S}_n(\gamma_l) \\
        =& \frac{\sum_{i=1}^M \alpha_i^2 l_i}{\sum_{i=1}^M \alpha_i^2} 
        = \sum_{i=1}^M \bar{\alpha}_i^2 l_i
    \end{split}
\end{equation}

Since $Y_n$ is positive semidefinite, $\expe Y_n$ is also positive semidefinite, and hence $\bar{A}_n = R_n^T \expe Y_n (R_n^{-1})^T$ is positive semidefinite as well. This means that $l_i \geq 0, \forall i \in \{1, ..., M \}$. Since $\tilde{L}_n(\tilde{\beta_g}) = \tilde{S}_n(\tilde{\gamma}_g) = \tilde{S}_n(\hat{\gamma}_1) = l_1$, there is
\begin{equation}
\label{bound_second_term_ito_cos}
    |\tilde{L}_n(\tilde{\beta_g}) - \tilde{L}_n(\beta_l)| \leq (1 - \bar{\alpha}_1^2) l_1 \leq (1 - \cos^2(\gamma_l, \tilde{\gamma}_g)) \lambda_1 (\bar{A}_n)
\end{equation}

Next, we bound the first and the third term on the right hand side of the inequality in Lemma \ref{break_into_three}.

\begin{lemma}
\label{perturbation_loss}
$\forall \beta$,
\begin{equation}
\label{bound_1_and_3}
    |L_n(\beta) - \tilde{L}_n(\beta)| \leq (\expe \| \Delta B_n \|^3)^{\frac{1}{3}} \cdot (\expe |\lambda_1(A_n)|^3)^{\frac{1}{3}} \cdot \left(\expe |\frac{1}{\lambda_M(B_n)}|^3\right)^{\frac{1}{3}}
\end{equation}
\end{lemma}

\begin{proof}[Proof of Lemma \ref{perturbation_loss}]
Let $\gamma = T_n^T \beta$.
\begin{equation}
\begin{split}
    |L_n(\beta) - \tilde{L}_n(\beta)| =& S_n(\gamma) - \tilde{S}_n(\gamma) \\
    =& \left|\expe \frac{(\gamma^T \Delta B_n \gamma)(\gamma^T A_n \gamma)}{(\gamma^T B_n \gamma)(\gamma^T \gamma)}\right| \\
    \leq& \expe \frac{\| \Delta B_n \| |\lambda_1(A_n)|}{|\lambda_M(B_n)|}
\end{split}
\end{equation}
Thus, we get the desired lemma by the generalized H{\"o}lder's inequality.

\end{proof}

Combining inequality \ref{bound_cos_2}, inequality \ref{bound_second_term_ito_cos} and Lemma \ref{perturbation_loss}, we get
\begin{equation}
\begin{split}
    |L_n(\beta_l) - L_n(\beta_g)| \leq & 2 (\expe \| \Delta B_n \|^3)^{\frac{1}{3}} \cdot (\expe |\lambda_1(A_n)|^3)^{\frac{1}{3}} \cdot \left(\expe \left|\frac{1}{\lambda_M(B_n)}\right|^3\right)^{\frac{1}{3}} + (1 - \cos^2(\gamma_l, \tilde{\gamma}_g)) \lambda_1 (\bar{A}_n) \\
    \leq & 2 \mu_n \nu_n \delta_n + \mu_n \nu_n \delta_n \eta_n C(\delta_n, \nu_n) \cdot \lambda_1 (\bar{A}_n)
\end{split}
\end{equation}

Meanwhile,
\begin{equation}
    \begin{split}
        |L_n(\beta_g) - \tilde{L}_n(\tilde{\beta}_g)| \leq & \max\{ |L_n(\beta_g) - \tilde{L}_n(\beta_g)|, |L_n(\tilde{\beta}_g) - \tilde{L}_n(\tilde{\beta}_g)| \} \\
        \leq & (\expe \| \Delta B_n \|^3)^{\frac{1}{3}} \cdot (\expe |\lambda_1(A_n)|^3)^{\frac{1}{3}} \cdot (\expe |\frac{1}{\lambda_M(B_n)}|^3)^{\frac{1}{3}} \\
        \leq & \mu_n \nu_n \delta_n
    \end{split}
\end{equation}
Hence, 
\begin{equation}
    \begin{split}
        L_n(\beta_g) \geq & \tilde{L}_n(\tilde{\beta}_g) - \mu_n \nu_n \delta_n \\
        \geq & \lambda_1(\bar{A}_n) - \mu_n \nu_n \delta_n \\
        \geq & \eta_n^{-1} - \mu_n \nu_n \delta_n 
    \end{split}
\end{equation}, or 
\begin{equation}
    \lambda_1(\bar{A}_n) \leq L_n(\beta_g) + \mu_n \nu_n \delta_n
\end{equation}

Therefore, 
\begin{equation}
    \begin{split}
        |L_n(\beta_l) - L_n(\beta_g)| \leq & 2 \mu_n \nu_n \delta_n + (1 - \cos^2(\gamma_l, \tilde{\gamma}_g)) [L_n(\beta_g) + \mu_n \nu_n \delta_n] \\
        \leq & \mu_n \nu_n \delta_n [2 + \eta_n \mu_n \nu_n \delta_n C(\delta_n, \nu_n)] + \eta_n \mu_n \nu_n \delta_n C(\delta_n, \nu_n) L_n(\beta_g) \\
        \leq & L_n(\beta_g) \left\{ \frac{\mu_n \nu_n \delta_n [2 + \eta_n \mu_n \nu_n \delta_n C(\delta_n, \nu_n)]}{\eta_n^{-1} - \mu_n \nu_n \delta_n} + \eta_n \mu_n \nu_n \delta_n C(\delta_n, \nu_n) \right\} \\
        =& \frac{2 \eta_n \mu_n \nu_n \delta_n [2 + C(\delta_n, \nu_n)]}{1 - \eta_n \mu_n \nu_n \delta_n} L_n(\beta_g)
    \end{split}
\end{equation}

Hence, we have proved the theorem, with $\epsilon_{\eta_n, \mu_n, \nu_n, \delta_n} = \frac{2 \eta_n \mu_n \nu_n \delta_n [2 + C(\delta_n, \nu_n)]}{1 - \eta_n \mu_n \nu_n \delta_n}$. $\square$

\section{Background}
\label{sec:background}

\subsection{Graph Min-Cuts and Spectral Clustering}
\label{laplaciansetup}

We consider graphs $G=(V,E)$, modeling a system 
of $N=|V|$ elements presumed to exhibit some form of community structure. 
The adjacency matrix $A$ associated with $G$ is the $N \times N$ binary matrix 
such that $A_{i,j} = 1$ when $(i,j) \in E$ and $0$ otherwise. We assume for simplicity
that the graphs are undirected, therefore having symmetric adjacency matrices.
The community structure is encoded in a discrete label vector $s : V \to \{1, \dots, C\}$
that assigns a community label to each node, and the goal is to 
estimate $s$ from observing the adjacency matrix. 

In the binary case, we can set $s(i) = \pm 1$ without loss of generality. Furthermore, we assume that the communities are associative, which means two nodes from the same community are more likely to be connected than two nodes from the opposite communities. The quantity
$$\sum_{i,j} (1 - s(i) s(j)) A_{i,j} $$
measures the cost associated with \emph{cutting} the graph between the two communities encoded by $s$, and we wish to minimize it
under appropriate constraints \citep{newman2006modularity}. Note that 
$\sum_{i,j} A_{i,j} = s^T D s$, with $D = \text{diag}(A {\bf 1})$ being the degree matrix, and so the cut cost can be expressed as a positive semidefinite quadratic 
form 
$$\min_{s(i) = \pm 1} s^T (D - A) s = s^T \Delta s$$
that we wish to minimize. 
This shows a fundamental connection between the community structure 
and the spectrum of the graph Laplacian $\Delta = D - A$, 
which provides a powerful and stable relaxation of the discrete combinatorial 
optimization problem of estimating the community labels for each node.
The eigenvector of $\Delta$
associated with the smallest eigenvalue is, trivially, $\mathds{1}$, but its Fiedler 
vector (the eigenvector associated with the second smallest eigenvalue) 
reveals important community information of the graph 
under appropriate conditions \citep{newman2006modularity}, and is associated with the graph conductance 
under certain normalization schemes \citep{spielman}. 


Given linear operator $\mathcal{L}(A)$ extracted from the graph (that we assume symmetric),
we are thus interested in extracting eigenvectors at the edge of its spectrum.
A particularly simple algorithm is the power iteration method. 
Indeed, the Fiedler vector of $\mathcal{L}(A)$ can be obtained by first extracting 
the leading eigenvector $v$ of $\tilde{A} = \|\mathcal{L}(A) \| \mathbb{I} - \mathcal{L}(A) $, and 
then iteratively compute
$$y^{(n)} = \tilde{A} w^{(n-1)}~~,~w^{(n)} = \frac{y^{(n)} - \langle y^{(n)}, v \rangle v}{\| y^{(n)} - \langle y^{(n)}, v \rangle v \| }~.$$
Unrolling power iterations and recasting the resulting model as a trainable neural network is akin to the LISTA sparse coding model, which unrolled iterative proximal splitting algorithms \citep{LISTA}. 

Despite the appeal of graph Laplacian spectral approaches, it is known
that these methods fail in sparsely connected graphs \citep{krzakala2013spectral} . Indeed, in such scenarios, the eigenvectors of the graph Laplacian concentrate on nodes with dominant degrees, losing their correlation with the community structure. In order to overcome this important limitation, people have resorted to ideas inspired from statistical physics, as explained next.

\subsection{Probabilistic Graphical Models and Belief-Propagation (BP)}
\label{bpsection}

Graphs with labels on nodes and edges can be cast as a graphical model where the aim of clustering is to optimize label agreement.  This can be seen as a posterior inference task. 
If we simply assume the graphical model is a Markov Random Field (MRF) with trivial compatibility functions for cliques greater than 2, the probability of a label configuration $\sigma$ is given by 
\begin{equation}
     \mathbb{P}(\sigma) = \frac{1}{\mathcal{Z}} \displaystyle \prod_{i \in V} \phi_i(\sigma_i) \displaystyle \prod_{ij \in E} \psi_{ij}(\sigma_i, \sigma_j).
\end{equation}

Generally, computing marginals of multivariate discrete distributions is exponentially hard.  For instance, if $X$ is the state space that we assume to be discrete, naively we have to sum over $|X|^{n-1}$ terms in order to compute $\mathbb{P}(\sigma_i)$. But if the graph is a tree, we can factorize the MRF efficiently to compute the marginals in linear time via a dynamic programming method called the sum-product algorithm, also known as belief propagation (BP).  An iteration of BP is given by 
\begin{equation}
\label{bpeqs}
     b_{i \rightarrow j}(\sigma_i) = \frac{1}{Z_{i\rightarrow j}} \phi_i(\sigma_i) \displaystyle \prod_{k \in \mathcal{N}(i)\setminus \{j\}} \displaystyle \sum_{\sigma_k \in X}\psi_{ik}(\sigma_i, \sigma_k) b_{k \rightarrow i}(\sigma_k).
\end{equation} 
When the graph is a tree, the BP equations above converge to a fixed point \citep{information2009mezard}. 
Moreover, if we define
\begin{equation}
    b_i(\sigma_i) = \prod_{k \in \mathcal{N}(i)} b_{k \rightarrow i}(\sigma_i)~,
\end{equation}
then at the fixed point, we can recover the single-variable marginals as $\mathbb{P}_i(\sigma_i) = b_i(\sigma_i)$.
For graphs that are not trees, BP is not guaranteed to converge.
However, graphs generated from SBM are locally tree-like so that such an approximation is reasonble \citep{abbe2017community}.

In order to apply BP for community detection, we need a generative model of the graph.  If the parameters of the underlying model are unknown, they parameters can be estimated using expectation maximization, which introduces further complexity and instability to the method since it is possible to learn parameters for which BP does not converge.



\subsection{Spectral method with the Non-backtracking and Bethe Hessian matrices}
\label{app.spectral}

The BP equations have a trivial fixed-point where every node takes equal probability in each group. Linearizing the BP equation around this point is equivalent to spectral clustering using the non-backtracking matrix (NB), a matrix defined on the directed edges of the graph that indicates whether two edges are adjacent and do not coincide. Spectral clustering using NB gives significant improvements over spectral clustering with different versions of the Laplacian matrix $L$ and the adjacency matrix $A$ \citep{krzakala2013spectral}. High degree fluctuations drown out the signal of the informative eigenvalues in the case of A and L, whereas the eigenvalues of NB are confined to a disk in the complex plane except for the eigenvalues that correspond to the eigenvectors that are correlated with the community structure, which are therefore distinguishable from the rest.   

However, spectral method with NB is still not optimal in that, firstly, NB is defined on the edge set; and secondly, NB is asymmetric, therefore unable to enjoy tools of numerical linear algebra for symmetric matrices. \cite{saade2014spectral} showed that a spectral method can do as well as BP in the sparse SBM regime using the Bethe Hessian matrix defined as $ BH(r):= (r^2-1) I - rA+D$, where $r$ is a scalar parameter.  This is thanks to a one-to-one correspondence between the fixed points of BP and the stationary points of the Bethe free energy (corresponding Gibbs energy of the Bethe approximation) \citep{saade2014spectral}. The Bethe Hessian is a scaling of the Hessian of the Bethe free energy at an extremum corresponding to the trivial fixed point of BP. Negative eigenvalues of $BH(r)$ correspond to phase transitions in the Ising model where new clusters become identifiable. 

The success of the spectral method using the Bethe Hessian gives a theoretical motivation for having a family of matrices including $I$, $D$ and $A$ in our GNN defined in Section \ref{gnnsec}, because in this way the GNN is capable of expressing the algorithm of performing power iteration on the Bethe Hessian matrices. 
On the other hand, while belief propagation requires a generative model and the spectral method using the Bethe Hessian requires the selection of the parameter $r$, whose optimal value also depends on the underlying generative model, the  GNN does not need a generative model and is able to learn and then make predictions in a data-driven fashion.

\subsection{Stochastic Block Model}
\label{sbmsec}
We briefly review the main properties needed in our analysis, and refer the interested reader to \cite{abbe2017community} for an excellent review.
The stochastic block model (SBM) is a generative model of random undirected graph denoted by $SBM(n,p,q, C)$. To generate a graph according to this model, one starts with a set $V$ of $n$ vertices each belong to one of $C$ communities, represented by a labeling function $F: V \rightarrow \{1, \dots, C\}$. We say the SBM is \textit{balanced} if the communities are of the same size.
Then, edges are constructed independently at random such that two vertices $u, v$ share an edge with probability $p$ if $F(v) = F(u)$, and with probability $q$ if $F(v) \neq F(u)$. 

The goal of community detection is then to estimate $F$ from the edge set. Let $\bar{F}:V \rightarrow \{1,C\}$ be the estimate of $F$ obtained by a certain method. For a sequence of $(p_n, q_n)$,
we say the method achieves \textit{exact recovery} if $\mathbb{P}( F_n = \bar{F_n})$ converges to $1$ as $n \to \infty$, and achieves \textit{weak recovery} or \textit{detection} if $\exists \epsilon > 0$ such that $\mathbb{P}(|F-\bar{F}| \geq 1 / C +\epsilon)$ converges to $1$ as $n \to \infty$ (i.e the method performs better than random guessing).


It is harder to tell communities apart if $p$ is close to $q$ (for example, if $p=q$ we get an Erd\H{o}s-Renyi random graph, which has no communities). In the binary case, it has been shown that exact recovery is possible on $SBM(n, (a \log n) / n, (b \log n) / n, 2)$ if and only if $(a+b) / 2 \geq 1 + \sqrt{ab}$ \citep{MNS,ABH}. While this is an information-theoretic result, it is also known that when this inequality is satisfied, there exist polynomial time algorithms that achieves exact recovery \citep{abbe2017community,MNS}. For this reason, we say that there is no \textit{information-computation gap} in this regime.

Note that for exact recovery to be possible, $p, q$ must grow at least $O(\log n)$ or else the graphs will likely not be connected, and thus the vertex labels will be underdetermined. Hence, in the sparser regime of constant degree, $SBM(n , a / n, b / n, C)$, detection is the best we could hope for. The constant degree regime is also of most interest to us for real world applications, as most large datasets have bounded degree and are extremely sparse. It is also a very challenging regime; spectral approaches using the adjacency matrix or the graph Laplacian in its various (un)normalized forms, as well as semidefinite programming (SDP) methods do not work well in this regime due to large fluctuations in the degree distribution that prevent eigenvectors from concentrating on the clusters \citep{abbe2017community}.
\cite{decelle2011asymptotic} first proposed the BP algorithm on the SBM, which was proven to yield Bayesian optimal values in \cite{coja2016information}. 
In the constant degree regime with $C$ balanced communities, the signal-to-noise ratio is defined as $SNR = (a-b)^2/(C(a+(C-1)b))$, and $SNR = 1$ is called the Kesten-Stigum (KS) threshold \citep{kesten1966limit}. 
When $SNR > 1$, detection can be solved in polynomial time by BP \citep{abbe2017community, decelle2011asymptotic}. For $C=2$, it has been shown that when $SNR < 1$, detection is information-theoretically unsolvable, and therefore $SNR = 1$ is both the computational and the information theoretic threshold \citep{abbe2017community}. For $C \geq 4$, it conjectured that for certain $SNR < 1$, there exist non-polynomial time algorithms able to solve the detection problem, while no polynomial time algorithm can solve detection when $SNR < 1$, in which case a gap would exist between the information theoretic threshold and the computational threshold \citep{abbe2017community}.


\section{Further experiments}
\subsection{Geometric Block Model}
\label{GBMsec}

\begin{table}[h]
\centering
\caption{\footnotesize{Performance measured by the overlap (in percentage) of GNN and LGNN on graphs generated by the Geometric Block Model compared with two spectral methods}}
\label{gbm_table}
{\small 
\begin{tabular}{|c|c|c|c|}
\hline
Model & $S=1$ & $S=2$  & $S=4$ \\
\hline 
Norm. Laplacian  & $1 \pm 0.5$ & $1 \pm 0.6$ & $1 \pm 1$ \\
Bethe Hessian & $18 \pm 1$ &  $38 \pm 1$ & $38 \pm 2$ \\
GNN & $20 \pm 0.4$  & $39 \pm 0.5$ & $39 \pm 0.5$ \\
LGNN & $\mathbf{22\pm 0.4}$  & $\mathbf{50 \pm 0.5}$ & $\mathbf{76 \pm 0.5}$ \\
\hline 
\end{tabular}
}
\end{table}

The success of BP on the SBM relies on its locally hyperbolic property, 
which makes the graph tree-like with high probability. This behavior is completely different if one 
considers random graphs with locally Euclidean geometry. 
The Geometric Block Model \citep{sankararaman2018community} is a random graph generated as follows. 
We start by sampling $n$ points $x_1,\dots, x_n$ i.i.d. from a Gaussian mixture model whose means are
$\mu_1, \dots \mu_k \in \mathbb{R}^d$ at distances $S$ apart and covariances are the identity. 
The label of each sampled point corresponds to which Gaussian it belongs to.
We then draw an edge between two nodes $i, j$ if $\| x_i - x_j\| \leq T/\sqrt{n}$. 
Due to the triangle inequality, the model contains a large number of short cycles, which affects 
the performance of BP \citep{information2009mezard}. This motivates other estimation algorithms 
based on motif-counting  that require knowledge of the model likelihood function \citep{sankararaman2018community}.

Table \ref{gbm_table} shows the performance of GNN and LGNN on the binary GBM model, 
obtained with $d=2$, $n=500$, $T=5 \sqrt{2}$ and varying $S$, compared to those of two spectral methods, 
using respectively the normalized Laplacian and the Bethe Hessian, with the latter introduced in Appendix \ref{app.spectral}.
We note that LGNN model, thanks to its added flexibility and the multiscale nature of its generators, is able to 
significantly outperform both spectral methods and the baseline GNN.

\subsection{Mixture of binary SBM}
\label{sec:furthersbm}
\begin{figure*}
     \centering
     \includegraphics[width=0.95\linewidth,height=2.4in]{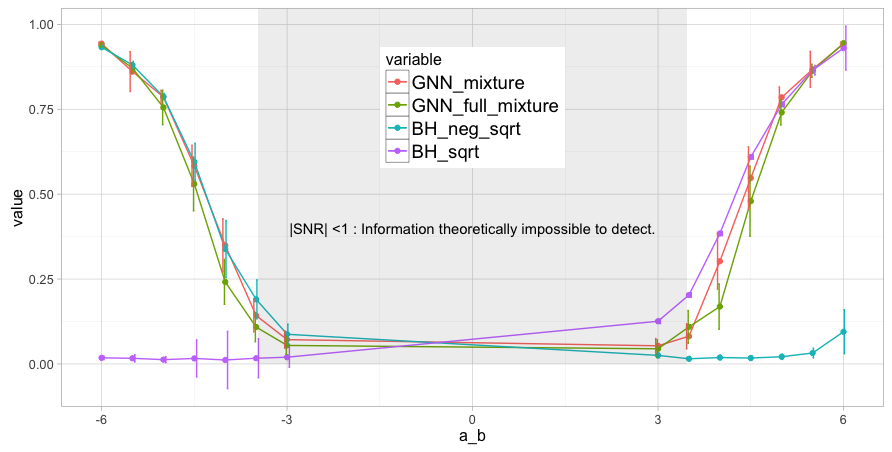}
     \caption{GNN mixture (Graph Neural Network trained on a mixture of SBM with average degree 3), GNN full mixture (GNN trained over different SNR regimes), $BH(\sqrt{\bar{d}})$ and $BH(-\sqrt{\bar{d}})$. {\it left: $k=2$}. We verify that $BH(r)$ models cannot perform detection at both ends of the spectrum simultaneously.}
     \label{fig:BH2}
\end{figure*}

We report here our experiments on the SBM mixture, generated with 
$$G \sim SBM(n=1000, p = k\bar{d} - q, q \sim \text{Unif}(0, \bar{d} - \sqrt{\bar{d}}), C=2)~,$$
where the average degree $\bar{d}$ is either fixed constant or also randomized with $\bar{d} \sim \text{Unif}(1, t)$. 
Figure \ref{fig:BH2} shows the overlap obtained by our model compared with several baselines. Our GNN model is either competitive with or outperforming the spectral method using Bethe Hessian (BH), which is the state-of-the-art spectral method \cite{saade2014spectral}, despite not having any access to the underlying generative model (especially in cases where GNN was trained on a mixture of SBM and thus must be able to generalize the $r$ parameter in BH). They all outperform by a wide margin spectral clustering methods using the symmetric Laplacian and power method applied to $\|BH\| I - BH$ using the same number of layers as our model. Thus GNN's ability to predict labels goes beyond approximating spectral decomposition via learning the optimal $r$ for $BH(r)$. The model architecture could allow it to learn a higher dimensional function of the optimal perturbation of the multiscale adjacency basis, as well as nonlinear power iterations, that amplify the informative signals in the spectrum. 

\subsection{Ablation studies of GNN and LGNN}
\label{ablations}
\begin{table}[h]

{\small 
\begin{tabular}{lllllllll}
 & & \#layers & \#features & J & $\mathcal{F}$ & \#parameters & Avg. & Std. Dev. \\
(a) & GNN & 30 & 8 & 2 & $I, D, A, A_2$ & 8621 & 0.1792 & 0.0385 \\
(b) & GNN & 30 & 8 & 4 & $I, D, A, ..., A_{4}$ & 12557 & 0.1855 & 0.0438 \\
(c) & GNN & 30 & 8 & 11 & $I, D, A, ..., A_{11}$ & 26333 & 0.1794 & 0.0359 \\
(d) & GNN & 30 & 15 & 2 & $I, D, A, A_{2}$ & 28760 & 0.1894 & 0.0388 \\
(e) & GNN & 30 & 12 & 4 & $I, D, A, ..., A^{(4)}$ & 27273 & 0.1765 & 0.0371 \\
(f) & LGNN & 30 & 8 & 2 & $I, D, A, A_{2}$ & 25482 & 0.2073 & 0.0481 \\
(g) & LGNN-L & 30 & 8 & 2 & $I, D, A, A_{2}$ & 25482 & 0.1822 & 0.0395 \\
(h) & LGNN & 30 & 8 & 2 & $I, A, A_{2}$ & 21502 & 0.1981 & 0.0529 \\
(i) & LGNN & 30 & 8 & 2 & $D, A, A_{2}$ & 21502 & 0.2212 & 0.0581 \\
(j) & LGNN & 30 & 8 & 2 & $A, A_{2}$ & 17622 & 0.1954 & 0.0441 \\
(k) & LGNN & 30 & 8 & 1 & $I, D, A$ & 21502 & 0.1673 & 0.0437 \\
(l) & LGNN-S & 30 & 8 & 2 & $I, D, A, A_{2}$ & 21530 & 0.1776 & 0.0398
\end{tabular}
  }
  \caption{\footnotesize{The effects of different architectures and choices of the operator family for GNN and LGNN, as demonstrated by their performance on the 5-class disassortative SBM experiments with the exact setup as in Section \ref{SBMsec2}. For LGNN, $\mathcal{F}'$ is the same as $\mathcal{F}$ except for changing $A$ to $B$.}}
  \label{ablations-table}
  \centering
\end{table}

Compared to $f$, each of $h$, $i$ and $k$ has one fewer operator in $\mathcal{F}$, and $j$ has two fewer. We see that with the absence of $A_{2}$, $k$ has much worse performance than the other four, indicating the importance of the power graph adjacency matrices. Interestingly, with the absence of $I$, $i$ actually has better average accuracy than $f$. One possibly explanation is that in SBM, each node has the same expected degree, and hence $I$ may be not very far from $D$, which might make having both $I$ and $D$ in the family redundant to some extent. 

Comparing GNN models $a$, $b$ and $c$, we see it is not the case that having larger $J$ will always lead to better performance. Compared to $f$, GNN models $c$, $d$ and $e$ have similar numbers of parameters but all achieve worse average test accuracy, indicating that the line graph structure is essential for the good performance of LGNN in this experiment. In addition, $l$ also performs worse than $f$, indicating the significance of the non-backtracking line graph compared to the symmetric line graph.

\end{document}



\maketitle

\begin{abstract}
\end{abstract}

\section{Further SBM Details}

\begin{definition}The \textit{Stochastic Block Model (SBM)} is a random graph model that can be defined by the following three parameters.
\begin{align*}
    n & : \text{ The number of vertices in the graph. }\\
    F & : V \rightarrow \{1, ..., k\} \\
    W & \in \mathbb{R}^{k \times k}_{\geq 0}: 
\end{align*}

where $F$ is a function on the vertex set that assigns a label to each vertex ($k$ is the number of communities) and $W$ is the symmetric matrix of probabilities of connections between the $k$ communities.

Given the above parameters, one can sample a SBM$(n, W)$ graph, call it $G = (V, E)$ (where $V$ is the vertex set and $E$ is the edge set, and $n= |V|$) by connecting two vertices $u, v \in V$ with probability $W_{ij}$, where $W_{ij}$ is the $ij^{th}$ entry of $W$, $v \in c_i$ and $u \in c_j$.  Whether one edge is in the SBM or not is independent of other edges and is solely determined by $W$ and $F$. We call a graph that is sampled in the above way an instantiation of the SBM$(n, W, F)$.
\end{definition}

In a simple case that we call the \textit{balanced SBM}, we have $k$ communities of the same size, and can specify the SBM model with three scalar parameters: $n$, $c_{in}, c_{out}, k$. The number of vertices is given by $n$, $c_{in}/n$ is the probability of connecting two vertices if they are from the same community and $c_{out}/n$ is the probability of connecting between two nodes of different communities

We say we can perform \textit{exact recovery} on a sequence of $\{ SBM(n, c_{in}, c_{out})\}_n$ if $\mathbb{P}( F_n = \bar{F_n}) \rightarrow_n 1.$  Where $F_n : V \rightarrow \{0,...,k\}$ corresponds to the correct clustering for SBM(n, p, q) and $\bar{F_n}: V \rightarrow \{0,...,k\}$ is the predicted cluster assignments. We we can perform \textit{detection} if the predicted labels correlate with the true labels. That means $\exists \epsilon > 0 : \mathbb{P}(|F_n-\bar{F_n}| \geq 1/k+\epsilon) \rightarrow_n 1.$

The detection regime is not just a weaker regime, it is actually impossible to have exact recovery for some families of $\{SBM(n, c_{in}, c_{out})\}_n$.  For instance when the SBM is not connected the isolated vertices would have underdetermined community membership.  It has been proven in the 2 community case that exact recovery is possible when the average degree grows at least at order $O(n \log n)$. That means $p=\frac{alog(n)}{n}$ and $q = \frac{b log(n)}{n}$.  Exact recovery occurs if and only if $\frac{a+b}{2} \geq 1 + \sqrt{ab}$ \cite{REMEMBER TO CITE}. We refer to this as the information theoretic threshold for exact recovery. On the other hand, if our average degree is constant in $n$, that is $p = \frac{a}{n}, q = \frac{b}{n}$ and we are in the hard regime of only being able to do detection. This is because the graph is not even connected as $n \rightarrow \infty$. Detection is possible if and only if $(a-b)^2 > 2(a+b)$. This lower bound gives the information theoretic threshold for detection.

There have been two algorithms successful in predicting the community labels, even down to the information theoretic threshold. The first is message passing based methods, in particular the Belief Propagation algorithm.  The algorithm is quadratic in the number of clusters and requires one to know the connectivity parameters of the SBM.  The second is the spectral method, in particular using the adjacency matrix or graph Laplacian, projecting the vectices into a subspace spanned by it's extreme eigenvectors (ones assossiated witht he largest or smallest eigenvalues) and doing k-means on this smaller dimensional representation.  Spectral methods have the clear advantage of being efficient and non parametric when dealing with more thatn two or an unknown number of communities.  Doing spectral clustering on the non-backtracting matrix can achieve detection even down to the information theoretic threshold.  The non backtracking matrix is dervied from the edges of the graph, so can be expensive if not in the constant degree regime.  Additionally it is also not symmetric, so a lot of the most efficient numerical algebra algorithms that leverages symmetric cannot be applied to it. A final very powerful graph that the spectral method can apply to to achieve detection in the information theoretic limit is the Bethe Hessian.  This graph is symmetric, and has been show in \cite{BH} to be just as good as the non-backtracking matrix both empirically and for analytic reasons.  It furthermore generalizes easily to weighted graphs, making it apply more flexibly to data.  

\begin{definition}
The \textit{Bethe Hessian} matrix is a deformed Laplacian matrix defined as $$ BH(r):= (r^2-1)\mathbb{I} - rA+D, $$
where $r$ is a scalar value.   
\end{definition}

Notice first that if $r=1$, spectral clustering on the BH reduces to spectral clustering on the unnormalized graph Laplacian.  The parameter $r$ can be interpreted as a regularizer.  The issue with using the Laplacian for spectral clustering in sparse regimes is that the second eigenvector, which is the one that correlates with the community labels in the two community case, because less separated from the other eigenvectors.  Choosing a value of $r$ that can amplify the signal of the second eigenvector is crucial.  It was shown in \cite{BH} that for the SBM, if one chooses $r_c := \sqrt{c}$ where $c$ is the average degree, then the informative eigenvalues will be negative.  This occurs when we use analyze the spectrum of $H(r_c)$ (assortative) and $H(-r_c)$ (dissortative).  

-By the Gershgoin circle theorem $\exists r > 0 $ such that the $BH(r)$ is positive definite. It is also symmetric, so we know that as we start with $r$ big enough such that $BH(r)$ is positive definite and then take $r \downarrow$ we will eventually get negative eigenvalues.  These eigenvalues are in fact the ones that are useful for us.  

-number of negative eigenvalues also shown to indicate number of clusters in SBM.  

The above choice of $r_c$ is critical, and actually relates back to the belief propagation algorithm, tying together nicely the spectral and message passing approaches. For a general graph $G$, the right choice of $r_c$ is actually $\sqrt{\rho(B)}$ where $\rho(B)$ is the spectral radius of the non-bactraking matrix of $G$.  

\begin{definition}
The \textit{Non Backtracking} matrix of graph $G$ is given by $$B_{i \rightarrow j, k \rightarrow l} = \delta_{jk} (1- \delta_{il}) $$
where $\delta$ is the Kronecker delta function.  
\end{definition}

All the eigenvalues of B that are not 1 or -1 are given by the $\det BH(r)$ so now we have reduced the problem of deriving $\rho(B)$ which depends on $B$ into one that requires knowledge of $BH(r)$.

-belief propagation is related to the BH because fixed points to the belief propagation are minima of the Bethe Hessian free energy. 

\begin{definition}
The \textit{Bethe free energy} is an approximation of the energy of the following pariwise Ising model defined on a graph $G$:

$$ P(\{x\}) = \frac{1}{Z}\exp \bigg( \sum_{(i,j) \in E} a \tanh (\frac{1}{r}) x_i x_j \bigg)$$

where $\{x_i\}_i$ are binary random variables indexed by the nodes of $n$ (one can regard the model as magnets sitting on the nodes of the graph that can interact with each other via the edges).  $r$ controls the strength of the interaction between nodes and is commonly referred to the as the temperature parameter in statistical physics. 
\end{definition}

 To study this model, the Bethe approximation approximates the mean $\mathbb{E}{x_i}$ and correlations $\mathbb{E}x_i x_j$ via two paramters $m_i$ and $\xi_{ij}$ respectively.  Using this one can write an energy equation for configurations of $\{x_i\})_n$ on the Ising model:

 $$ F_{Bethe}(\{m_i\}, \{\xi_{ij}\}) = -\sum_{(i,j) \in E} a \tanh (\frac{1}{r}) \xi_{ij} + \sum_{(i, j) \in E}\sum_{x_i, x_j} \nabla \frac{1 + m_i x_i + m_j x_j + \xi_{ij} x_i x_j }{4}$$ 
 $$ + \sum_{i \in V} (1-d_i) \sum_{x_i} \nabla (\frac{a + m_i x_i}{2})$$

It was shown in \footnote{http://www.cs.princeton.edu/courses/archive/spr06/cos598C/papers/YedidaFreemanWeiss2004.pdf} that the fixed points of BP are the minima of the Bethe free energy.  Relating back to the BH, the Hessian of the Bethe free energy at the point $m_{ij} = 0, \xi_{ij} = \frac{1}{r}$, which is important because it is a stationary point of the Bethe free energy for every $r$, is in fact equal to $\frac{BH(r)}{r^2 -1}$ (hence the name Bethe Hessian).  Negative eigenvalues of $BH(r)$ correspond to phase transitions in the Ising model where new cluster becomes identifiable.  

\begin{figure}
\begin{center}
  \includegraphics[width=0.6\textwidth]{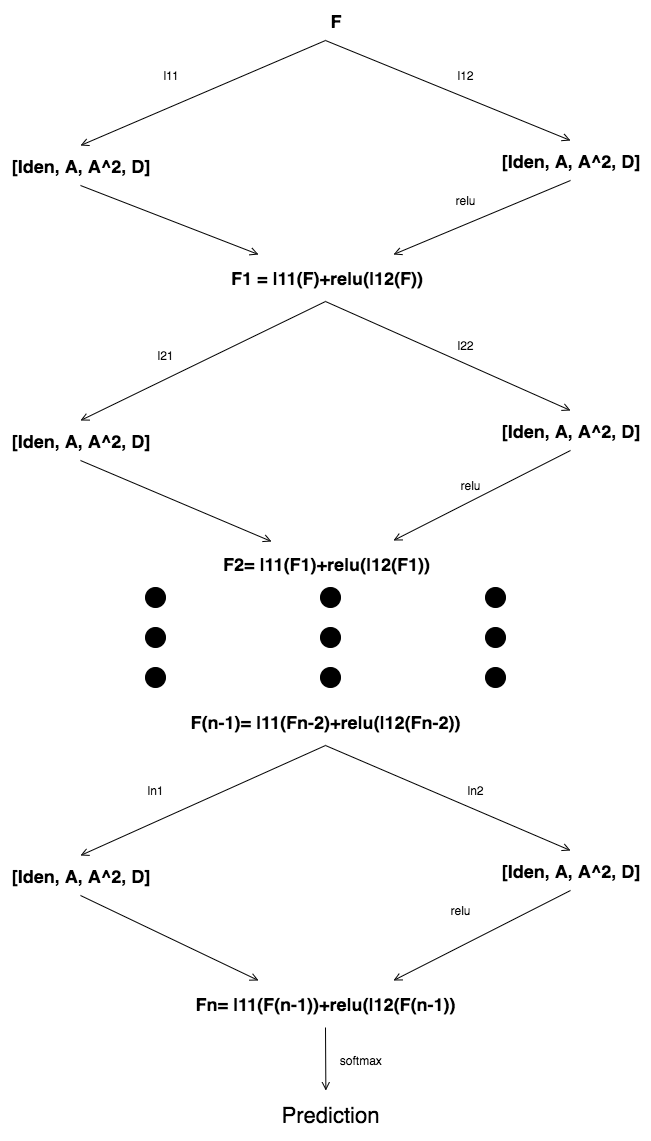}
  \caption{An example of an architecture constructed from operators D, W and $\theta$}
  \label{fig:GNN}
 \end{center}
\end{figure}

 \textbf{Interesting but can remove if no space in writeup:} \textit{Some intution for why is that SBMs above KS threshold enjoy an efficient recovery procedure because the uniform solution (guessing 1/k for all points) is highly unstable so the basin of attraction for the correct solution is basically the entire space of possible initializations for BP.  Whereas between ITT and KS, we have that uniform solution and the correct solution are both locally stable fixed points, however if messages (for BP) are intialized randomly we cannot do better than chance because the basin of attraction for the correct fixed point (correct community labels) becomes exponentially small  (thus attracting an exponentially small fraction of intializations).  So if one initializes close to the true predictions one will get the right answer, but otherwise no}

\section{Reproduced Figures}

For convenience, we reproduce the figures in higher resolution.

\begin{figure}
    \centering
    \includegraphics[width=0.5\linewidth]{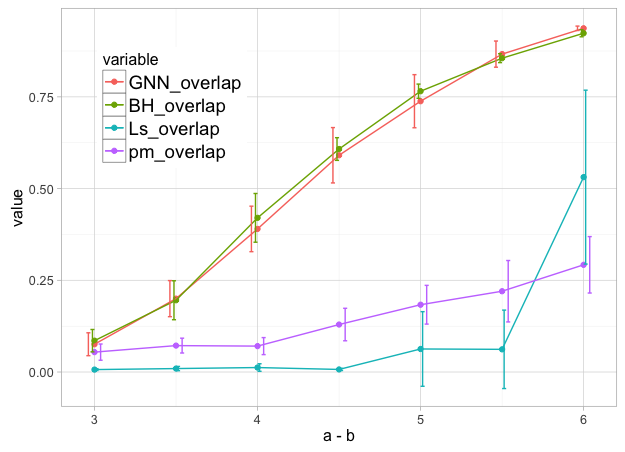}
    \includegraphics[width=0.5\linewidth]{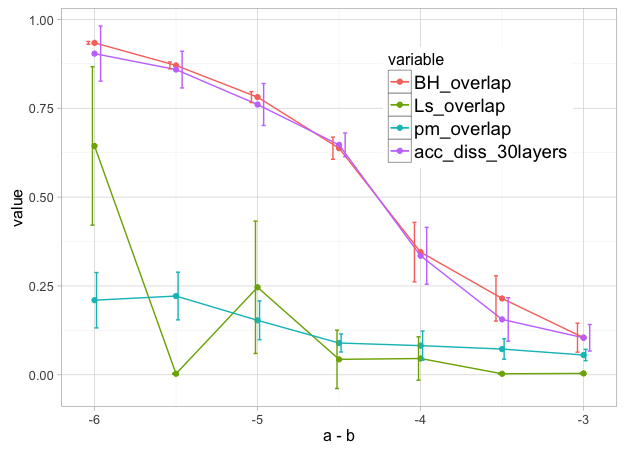}
    \caption{left: $k=2$ associative, right: $k=2$ disasocciative.  X-axis corresponds to 
    SNR , Y-axis is overlap; (see text)}
    \label{fig:BH2a}
\end{figure}

\begin{figure}
    \centering
    \includegraphics[width=0.5\linewidth]{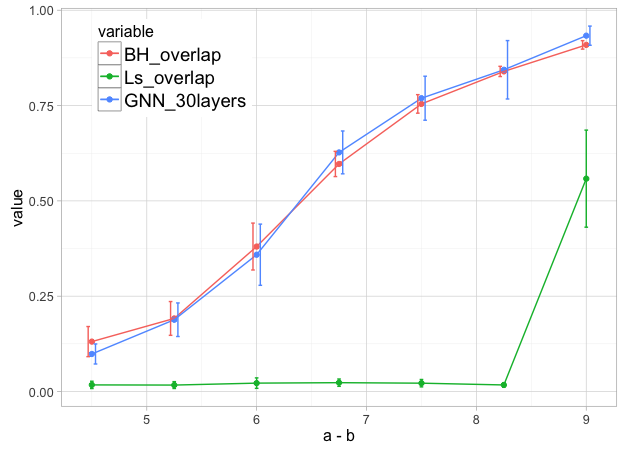}
    \includegraphics[width=0.5\linewidth]{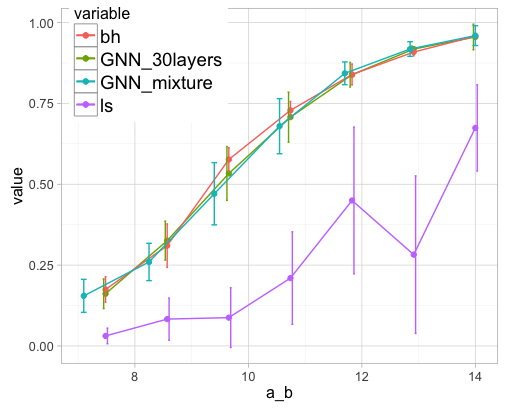}
    \caption{left: $k=3$ associative, right: $k=4$ associative. X-axis corresponds to 
    SNR , Y-axis is overlap; (see text)}
    \label{fig:BH3a}
\end{figure}

\begin{figure}
     \centering
     \includegraphics[width=1\linewidth]{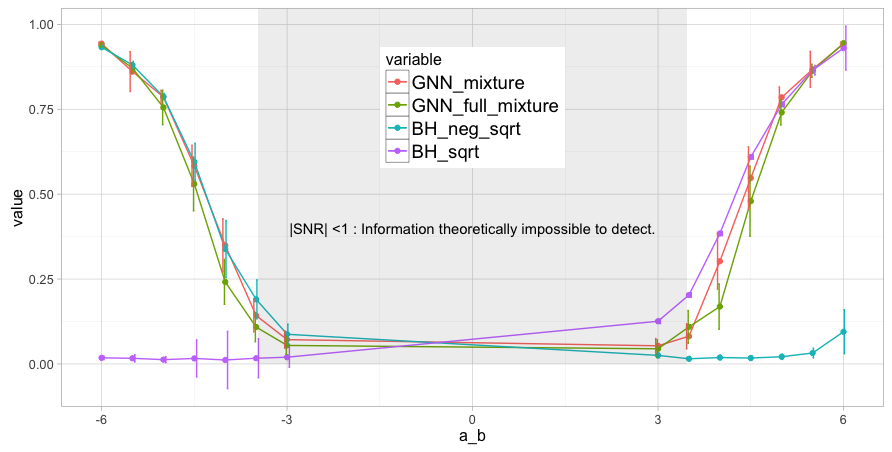}
     \caption{GNN mixture (Graph Neural Network trained on a mixture of SBM with average degree 3), GNN full mixture (GNN trained over different SNR regimes some below threshold), $BH(\sqrt{\bar{d}})$ and $BH(-\sqrt{\bar{d}})$. We verify that Bethe Hessian models cannot perform detection at both ends of the spectrum simultaneously.}
     \label{fig:BH4a}
\end{figure}





\printbibliography
